\PassOptionsToPackage{table,dvipsnames}{xcolor}
\documentclass{applemlr}

\usepackage{amsmath}
\usepackage{enumerate}
\usepackage{algorithm}
\usepackage{algpseudocode}
\usepackage{amsfonts}
\usepackage{amsthm}
\usepackage{cleveref}
\usepackage{diagbox}
\usepackage{colortbl}
\usepackage{amssymb}
\usepackage{xspace}
\usepackage{wrapfig}
\usepackage{adjustbox}
\usepackage{tabularx}
\usepackage{longtable}
\usepackage{booktabs}
\usepackage{mathtools}
\usepackage{tikz}
\usepackage{enumitem}
\usepackage{silence}
\usepackage{dsfont}
\usepackage[table]{xcolor}
\usepackage[dvipsnames]{xcolor}
\usepackage{multirow}
\usepackage{makecell}
\usepackage{xfakebold}
\usepackage{amsmath,amsfonts,bm}

\def\eqref#1{equation~\ref{#1}}

\def\1{\bm{1}}

\DeclareMathAlphabet{\mathsfit}{\encodingdefault}{\sfdefault}{m}{sl}
\SetMathAlphabet{\mathsfit}{bold}{\encodingdefault}{\sfdefault}{bx}{n}

\usepackage{nicefrac}

\definecolor{textgray}{HTML}{6E6E73}
\usetikzlibrary{positioning, calc}
\usetikzlibrary{decorations.pathmorphing}
\usetikzlibrary{arrows.meta, decorations.pathreplacing, backgrounds}

\definecolor{correctgreen}{RGB}{0,145,65}
\definecolor{wrongred}{RGB}{175,30,30}
\definecolor{alignblue}{RGB}{55,120,195}
\definecolor{misalignred}{RGB}{195,55,55}
\definecolor{neutralgray}{RGB}{140,140,140}
\definecolor{highlightyellow}{RGB}{255,252,210}

\definecolor{mechanismFrozen}{HTML}{F5F6F4}
\definecolor{mechanismICL}{HTML}{FFF4C8}
\definecolor{mechanismSDPO}{HTML}{FBE9E8}
\definecolor{mechanismGRPO}{HTML}{EAF2FF}
\definecolor{mechanismICLSDPO}{HTML}{FFF0DF}
\definecolor{mechanismICLGRPO}{HTML}{ECFBF5}

\definecolor{systembg}{HTML}{F0F4FF}
\definecolor{systemframe}{HTML}{6B7FD7}
\definecolor{userbg}{HTML}{F6FFF6}
\definecolor{userframe}{HTML}{4CAF50}
\definecolor{demobg}{HTML}{FFF8F0}
\definecolor{demoframe}{HTML}{E8913A}
\definecolor{promptgray}{HTML}{F7F7F8}
\definecolor{promptborder}{HTML}{D1D5DB}
\definecolor{accentblue}{HTML}{3B82F6}
\definecolor{promptgreen}{HTML}{2E7D32}
\definecolor{promptred}{HTML}{C62828}
\definecolor{summcolor}{HTML}{E65100}
\definecolor{lessoncolor}{HTML}{00796B}

\makeatletter
\patchcmd{\wrong@fontshape}{\@gobbletwo}{}{}{}
\makeatother
\numberwithin{equation}{section}
\makeatletter
\AtBeginDocument{
  \urlstyle{sf}
  
}
\makeatother

\definecolor{light}{RGB}{125, 125, 125}
\crefname{tcb@cnt@pbox}{code}{code}
\Crefname{tcb@cnt@pbox}{Code}{Code}
\crefname{assumption}{assumption}{assumption}
\Crefname{assumption}{Assumption}{Assumptions}

\newtcolorbox[auto counter]{pbox}[2][]{
  colback=white,
  title=Code~\thetcbcounter: #2,
  #1,fonttitle=\sffamily,
  fontupper=\sffamily,
  arc=2pt,
  colframe=bgcolor,
  coltitle=fgcolor,
  colbacktitle=bgcolor,
  toptitle=0.25cm,
  bottomtitle=0.125cm
}

\makeatletter
\newcommand\applefootnote[1]{%
  \begingroup
  \renewcommand\thefootnote{}%
  \renewcommand\@makefntext[1]{\noindent##1}%
  \footnote{#1}%
  \addtocounter{footnote}{-1}%
  \endgroup
}
\makeatother

\definecolor{cverbbg}{gray}{0.90}

\usepackage{url}
\usepackage{xcolor}
\usepackage{booktabs}
\usepackage{array}
\usepackage{algorithm}
\usepackage{algpseudocode}
\usepackage{tikz}
\usetikzlibrary{arrows.meta,shadows,positioning}
\usepackage{graphicx}
\usepackage{placeins}
\usepackage{needspace}

\DeclareRobustCommand{\world}[1]{\textnormal{\scshape #1}}
\DeclareRobustCommand{\policy}[1]{\textnormal{\sffamily #1}}
\usepackage[most]{tcolorbox}
\definecolor{rqink}{HTML}{26466F}
\definecolor{rqtint}{HTML}{EFF3F8}
\def\rqstyle{3}
\newcommand{\researchquestion}[1]{%
  \par\vspace{0.3\baselineskip}%
  \ifcase\rqstyle
    \noindent\parbox{\linewidth}{\centering\bfseries #1}\par
  \or
    \begin{tcolorbox}[enhanced, colback=rqtint, colframe=rqink!55, boxrule=0.5pt,
      arc=3pt, left=10pt, right=10pt, top=5pt, bottom=5pt, halign=center,
      before skip=3pt, after skip=6pt]
      #1
    \end{tcolorbox}
  \or
    \begin{tcolorbox}[enhanced, frame hidden, colback=rqtint, sharp corners,
      borderline west={2.5pt}{0pt}{rqink}, left=12pt, right=10pt, top=5pt, bottom=5pt]
      #1
    \end{tcolorbox}
  \or
    \begin{tcolorbox}[enhanced, colback=white, colframe=rqink!70, boxrule=0.5pt, arc=3pt,
      attach boxed title to top center={yshift=-2mm}, title={Research questions},
      fonttitle=\small\scshape, coltitle=rqink,
      boxed title style={colback=white, colframe=white, boxrule=0pt},
      left=10pt, right=10pt, top=7pt, bottom=5pt, halign=center,
      before skip=8pt, after skip=6pt]
      #1
    \end{tcolorbox}
  \fi
  \vspace{0.3\baselineskip}}

\title{Where Should LLM Agents Adapt \\ at Test Time? In Context, in Weights, or Both}

\author[1,*]{Parsa~Mirtaheri}
\author[2]{Mohammadreza~Armandpour}
\author[2]{Keivan~Alizadeh}
\author[2]{Iman~Mirzadeh}
\author[2]{Parshin~Shojaee}
\author[2]{Aryo~Lotfi}
\author[2]{David~Harrison}
\author[2]{Mehrdad~Farajtabar}

\affiliation[1]{UC San Diego}
\affiliation[2]{Apple}
\contribution[*]{Work done at Apple.}

\abstract{
LLM agents operating repeatedly in the same environment, such as a code
repository, a database, or an application suite, encounter tasks that share
latent structure. Experience can reveal this structure, allowing agents to
reuse environment knowledge and learn from environmental feedback. Agents can
adapt at test-time using this experience through two channels: retaining
it in context and updating their weights. These channels are usually
studied separately, leaving unclear what each contributes and when they
complement each other. We compare in-context learning (ICL), reinforcement
learning with verifiable rewards using GRPO, and self-distillation using SDPO,
alone and in combination, through a shared online protocol and modular
evaluation framework. Agents encounter matched task streams, are evaluated on
each task before adapting to it, and learn only from their own attempts and
environment feedback. We evaluate 14 adaptation mechanisms and a stateless
baseline with Qwen3.5-9B on 109 tasks from four environments drawn from
SWE-bench Verified, AppWorld, ScienceWorld, and BIRD-Critic-SQLite, using five
task orders per environment. ICL provides the largest standalone gain: its
best policy raises mean success across environments from 28.8\% to 49.4\%.
Adding GRPO to this policy achieves the highest mean success, 51.6\%.
While adding GRPO improves the overall mean under every ICL policy, SDPO has
mixed effects. Trace analyses show fewer lookups and shorter attempts under ICL,
consistent with reusing knowledge of the shared structure. However, ICL also
frequently skips inspection and verification for the current task. Adding GRPO
to ICL helps restore these behaviors while preserving its reduced environment
lookups. These observations suggest that reinforcement learning with
verifiable rewards can refine how agents use retained experience,
motivating the joint study of context management and weight updates.}

\date{\sffamily\today}

\begin{document}

\maketitle

\section{Introduction}
\label{sec:intro}

\begin{figure}[t]
\centering
\includegraphics[width=\linewidth]{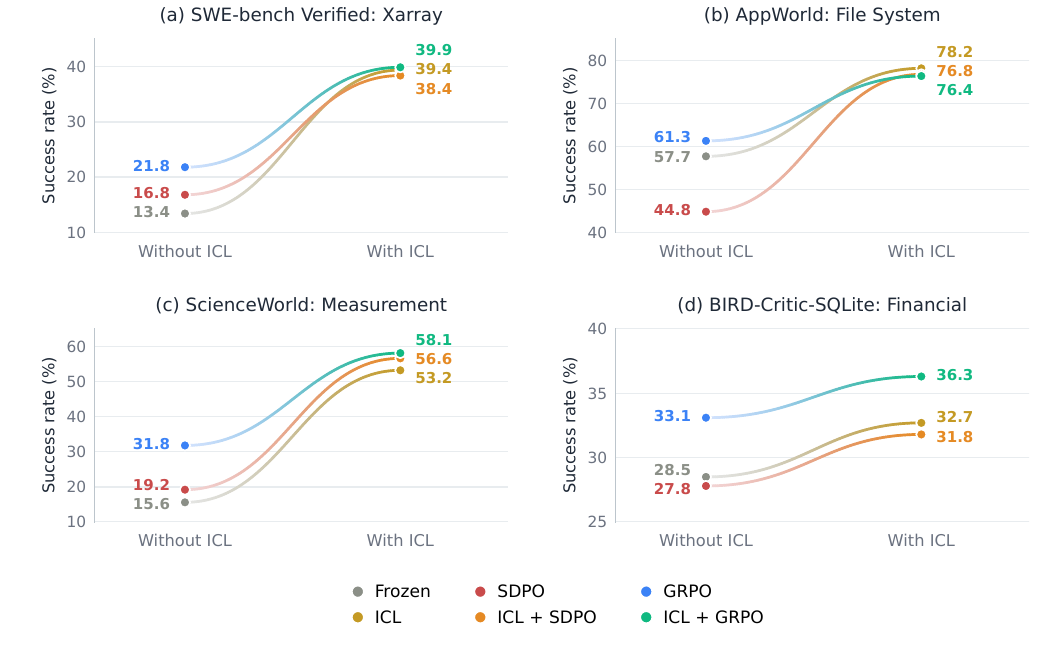}
\caption{\textbf{On average across benchmarks, in-context learning (ICL) gives the largest standalone gain, while ICL with GRPO gives the highest overall gain.} Points show five-seed mean success rates (\%), using the best mechanism within each adaptation family for each world. Curves connect results without and with ICL for the same weight-update policy. Standard errors are reported in Table~\ref{tab:main}.  }
\label{fig:intro-overview}
\end{figure}

Large language model (LLM) agents are increasingly deployed in persistent roles:
a coding agent working on one repository, an analyst querying one database, or an
assistant operating within one suite of applications
\citep{ouyang_reasoningbank_2026,wei_evo-memory_2026}. Tasks within each environment
share latent structure specific to it: the idioms of its codebase, the conventions
of its schema, or the behavior of its interfaces. By discovering this structure
through interaction and environmental feedback, an agent can reuse what it learns
on later tasks. For example, learning how a database is organized can let it answer
subsequent questions with far fewer queries \citep{asawa_continual_2026}.
How should the agent carry this experience forward? For a fixed architecture,
the LLM's behavior is governed by a policy $\pi_\theta(\cdot \mid c)$, giving
experience two routes to influence future behavior: through the context $c$
or the weights $\theta$ \citep{allard_experiential_2026,ouyang_reasoningbank_2026,fallah_claas_2026}. This paper asks where an agent should adapt at test time:
in context, in weights, or in both.

\emph{Test-time adaptation} lets an agent specialize to its own deployment environment while the base model stays unchanged for other deployments \citep{tandon_end--end_nodate,hubotter_specialization_2026}.
It has three defining properties. Every arriving task counts toward the agent's performance and also supplies experience for later tasks, so the agent is evaluated while it learns.
Accumulated state persists across the stream rather than resetting between tasks \citep{akyurek_surprising_2025,tandon_end--end_nodate}.
And the agent learns only from its own attempts and environmental feedback, without demonstrations or a stronger teacher \citep{silver_welcome_2025,hubotter_reinforcement_2026}.

Context and weights have complementary properties as adaptation channels \citep{tiwari_learning_2026}.
Adding experience to the context requires no training, so the agent can use it on the next task while its weights stay intact. However, adaptation through context is bounded by the context window, in which different experiences compete, and conditioning the model through it comes with an extra computational cost
\citep{hu_when_2026}. Weights persist without occupying context, but they are costly
to update, and an update that helps one task can hurt others
\citep{tiwari_learning_2026,luo_empirical_2025}. \citet{tandon_end--end_nodate}
interpret weights updated at test time as long-term memory and the context
window as short-term memory.

Methods for both channels are well developed. Memory-based agents
keep their weights frozen and retrieve reflections, workflows, cases, or strategies
from past attempts into their context
\citep{shinn_reflexion_2023,wang_agent_2024,zhou_memento_2025,zhang_agentic_2025,ouyang_reasoningbank_2026},
whereas test-time training methods write
experience into the weights
\citep{akyurek_surprising_2025,yuksekgonul_learning_2026,hubotter_reinforcement_2026}.
But the two lines of work have developed mostly apart
\citep{jiang_adaptation_2026,hebbar_sia_2026}, and differences in their tasks and
evaluation protocols make their reported gains difficult to compare directly.

Existing systems that optimize a prompt or harness jointly with model weights
suggest that the two adaptation channels can complement each other
\citep{dorovatas_modular_2026,tiwari_learning_2026,hebbar_sia_2026,kim_whale_2026}.
However, these results leave unclear what each channel contributes when agents
learn from their own interactions across tasks in a shared environment. We
investigate this through a controlled comparison addressing the following questions:

\researchquestion{What do context and weights each contribute to test-time adaptation?
Does combining them help? And how should the two be coordinated?}

We conduct extensive experiments using our framework, in which tasks from an agentic benchmark that share latent structure are arranged into a stream, and every adaptation mechanism processes the stream and is scored on each task as it arrives. The agent learns only from its own attempts and the feedback the
environment returns, and it may manage its context, weights, or both from one task to
the next. We build four \emph{worlds}, each a set of tasks that share one environment: issues for one repository of SWE-bench Verified \citep{jimenez_swe-bench_2024}, tasks using one application suite of AppWorld
\citep{trivedi_appworld_2024}, the measurement tasks of ScienceWorld
\citep{wang_scienceworld_2022}, and tasks on one database of BIRD-Critic-SQLite \citep{li_swe_sql_2025}.

We compare raw-trajectory ICL, reinforcement learning with verifiable rewards
using GRPO \citep{shao_deepseekmath_2024}, and self-distillation using SDPO
\citep{hubotter_reinforcement_2026}, alone and in combination. Raw-trajectory
replay lets us study context adaptation without introducing additional
summarization, retrieval, or judgment procedures, while keeping the source of
experience consistent across mechanisms. Prior findings also support its strength as a baseline by showing that raw
replay can outperform dedicated memory architectures
\citep{asawa_continual_2026}. We measure each mechanism's gain over the same agent run
without adaptation.

Our contributions are threefold:
{\setlength{\leftmargini}{1.2em}
\begin{itemize}
  \setlength{\itemsep}{2pt}\setlength{\parskip}{0pt}
  \item \textbf{Controlled study of adaptation mechanisms.} Under one
  protocol, which uses matched task streams, per-attempt limits, and a common
  source of environmental feedback
  (Section~\ref{sec:setup}), we compare a stateless baseline and adaptation
  mechanisms that adapt the context, the weights, or both, with Qwen3.5-9B on 109
  tasks from four benchmarks, each under five task orders (Section~\ref{sec:experiments}).
  \item \textbf{Analysis of when context and weights complement each other.}
  We show that ICL provides the largest standalone gain; adding GRPO improves
  the overall mean under every ICL policy, while SDPO has mixed effects
  (Figure~\ref{fig:intro-overview}; Table~\ref{tab:main}). Our trace analyses
  show that ICL reduces environment lookups but frequently skips task-specific
  inspection and verification. Adding GRPO helps restore these behaviors while
  preserving reduced lookups, suggesting that reinforcement learning with
  verifiable rewards can refine how agents use retained experience
  (Sections~\ref{sec:single-channel}--\ref{sec:hybrid-mechanisms}).
  \item \textbf{Extensible framework for evaluating adaptation mechanisms.} We provide a framework built on verl
  \citep{sheng_hybridflow_2024}, which registers benchmarks and adaptation
  mechanisms as modules, where one
  mechanism can combine in-context experience,
  reinforcement learning with verifiable rewards, and self-distillation.
\end{itemize}}

\section{Related Work}
\label{sec:related}

\paragraph{Adaptation through context.}
Agents retain experience as interaction histories \citep{park_generative_2023},
reflections \citep{shinn_reflexion_2023,zhao_expel_2024}, reusable workflows
\citep{wang_voyager_2023,wang_agent_2024}, or strategies and playbooks
\citep{ouyang_reasoningbank_2026,zhang_agentic_2025}. Experience selection can itself be
learned \citep{zhou_memento_2025}. These methods change the context while keeping
the model fixed. Raw history is already a strong baseline in CL-Bench
\citep{asawa_continual_2026}, but bounded context makes old and new experience
compete \citep{hu_when_2026}. We study raw-trajectory ICL policies to separate the
contribution of in-context adaptation itself from that of additional summarization and retrieval machinery.

\paragraph{Adaptation through weights.}
Test-time training adapts a model using the test input \citep{sun_test-time_2020}.
For LLMs, proposed methods fine-tune on retrieved neighbors of an input
\citep{hardt_test-time_2024}, on a task's few-shot examples
\citep{akyurek_surprising_2025}, or on synthetic variants of inputs an agent is
uncertain about \citep{acikgoz_self-improving_2025}. They also reinforce solutions to a
target problem \citep{yuksekgonul_learning_2026}, practice on related problems
selected for the target \citep{hubotter_learning_2025}, or train on an unlabeled test
set with majority-vote rewards \citep{zuo_ttrl_2025}. These methods mainly train for
one input or task, or for a fixed set known in advance, and then reset. Online
variants keep updating over a stream of inputs from one distribution, with
self-supervised losses
\citep{krause_dynamic_2018,sun_test-time_2020,hu_test-time_2025}. Although
reinforcement learning \citep{shao_deepseekmath_2024} and self-distillation
\citep{hubotter_reinforcement_2026} are mainly applied before deployment, recent
systems also apply them during deployment, updating persistent weights over a stream
of scenarios \citep{fallah_claas_2026} or agent tasks \citep{lu_ascent_2026}. Here, we
adopt both for continual adaptation: the adapter and optimizer state persist across a
deployment. This post-deployment continual adaptation setting differs from
traditional continual learning, which is centrally concerned with retaining general
capability under a changing task distribution
\citep{shi_continual_2024,luo_empirical_2025}.

\paragraph{Combining context and weights.}
A growing line of work adapts context and weights together, on the view that the two
channels are complementary \citep{dorovatas_modular_2026}. One family moves experience from
context into weights: context distillation transfers context-conditioned behavior into weights
\citep{snell_learning_2022,ye_-policy_2026}, online experiential learning alternates extracting
experience with consolidating it \citep{ye_online_2026}, and LifeSkill internalizes skills
during test-time interaction by reinforcing skill-conditioned trajectories
\citep{mao_learning_2026}. Another adapts the two side by side, jointly adapting prompts and
weights \citep{soylu_fine-tuning_2024,tiwari_learning_2026}, choosing between harness and
weight updates \citep{hebbar_sia_2026}, or using retained experience during reinforcement
learning \citep{wu_evolver_2026,liu_exploratory_2026}. How much each channel contributes,
however, remains unclear, because comparisons point in different directions. On fixed training
data, prompt optimization can outperform reinforcement learning \citep{agrawal_gepa_2026},
in-context learning can generalize more flexibly than fine-tuning
\citep{lampinen_generalization_2025}, and WHALE, which alternates harness search with
fine-tuning, finds that the limiting channel depends on the domain \citep{kim_whale_2026}. In
contrast, an online comparison on an adversarial instruction-following task finds that weight
updates outperform in-context learning \citep{fallah_claas_2026}. We compare context, weights,
and their combinations directly, online and in four agentic environments, under controlled
conditions (one protocol, the same model, the same task streams, and the same source of
feedback), to isolate each channel's contribution and test when combining them helps.

\paragraph{Evaluation and infrastructure.}
Recent benchmarks evaluate agents over sequential experience
\citep{wu_streambench_2024,wei_evo-memory_2026,he_memoryarena_2026}. CL-Bench
measures improvement over the same model without accumulated state
\citep{asawa_continual_2026}; we adopt this comparison. Closest to our setting,
\citet{harrington_when_2026} compare prompt-based, supervised, and
reinforcement-learning methods under one protocol, but their methods update between
stages on curated data, and none combines explicit in-context experience with weight
updates. Our agents, by contrast, learn only from their own attempts and environment
feedback. On the systems side, Agent Lightning trains arbitrary agents with
reinforcement learning \citep{luo_agent_lightning_2025}, EasyOPD unifies on-policy
distillation methods on a shared backend \citep{sun_easyopd_2026}, and CLaaS updates a
deployed agent's weights from its experience \citep{fallah_claas_2026}. Our framework
builds on verl \citep{sheng_hybridflow_2024} and runs context and weight adaptation in
a single loop. More complex adaptation methods can be added through its mechanism
interface and compared under the same protocol.

\section{Setup}
\label{sec:setup}

\begin{figure}[t]
\centering
\includegraphics[width=\linewidth]{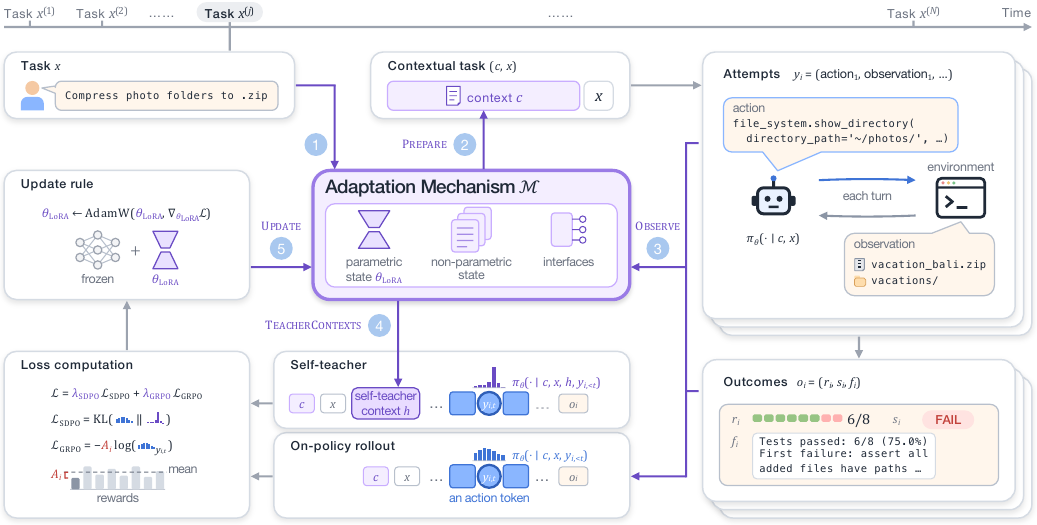}
\caption{One round on a task. (1)~The adaptation mechanism receives a task. (2)~\textsc{Prepare} returns a context $c$, with which the agent makes $k$ attempts $y_i$ in fresh sessions of the
environment. (3)~\textsc{Observe} receives the attempts and their outcomes $o_i$ and updates the mechanism's non-parametric state.
(4)~\textsc{TeacherContexts} adds hindsight $h$ for SDPO's self-teacher. (5)~The mechanism updates its parametric state using the weighted SDPO and GRPO losses.}
\label{fig:protocol-lifecycle}
\end{figure}

\subsection{Stream of related tasks}
\label{sec:setting}

Agentic benchmarks typically score each task on its own, with a fresh agent, yet many are
built on a stateful environment, such as a code repository, whose tasks share latent structure specific to it.
Selecting the tasks that share one environment and ordering them as a stream
recasts such a benchmark for test-time adaptation.
We define a \emph{world} $\mathcal{W}$ as a set of tasks
$\mathcal{X}=\{x_1,\dots,x_N\}$ that share an environment. A seed fixes the
\emph{stream} $(x^{(1)},\dots,x^{(N)})$, a random order in which each task appears
once. For each task, the agent makes $m$ rounds of $k$ parallel
\emph{attempts} (Figure~\ref{fig:protocol-lifecycle}). An attempt $y$ is a
multi-turn interaction with a fresh session of the environment that ends when the
agent submits a solution or reaches the goal state, or at a limit. The benchmark then
returns an \emph{outcome} $o=(r,s,f)$: a reward $r\in[0,1]$, a success indicator
$s\in\{0,1\}$, and textual feedback $f$, such as how many tests passed.
The $k$ attempts of a round share one context and one set of weights, so together
they estimate the agent's success rate and give the adaptation mechanism several outcomes
to compare. The mechanism adapts after every round, and each later round makes fresh
attempts with the adapted context and weights, adding experience from the same task. Only
the first round is scored, before any adaptation to the task (Section~\ref{sec:protocol}).
Table~\ref{tab:notation} summarizes the notation.

We derive four worlds from well-established benchmarks. In \world{Xarray}, the agent
resolves SWE-bench Verified issues of \texttt{pydata/xarray} with shell commands,
judged by each issue's tests (22 tasks); in \world{File System}, it writes Python that
calls the APIs of AppWorld's apps (21 tasks whose primary app is the file
system); in \world{Measurement}, it issues text commands in ScienceWorld's simulated
laboratory to measure a temperature or a melting point (22 tasks); and in
\world{Financial}, it fixes faulty SQL for one BIRD-Critic-SQLite database, which it
can query before submitting (44 tasks).
In each world, an existing grader scores every attempt, and the stateless baseline
leaves headroom (Table~\ref{tab:main}); gains over that baseline then test whether earlier
tasks inform later ones. Because our framework reuses each benchmark's tasks, environments,
and graders, other benchmarks whose tasks share an environment can be added the same way,
without building new tasks.

\begin{table}[b]
\caption{\textbf{Notation for the online protocol.} Indices $j$, $\ell$, and $i$
denote task position, round, and attempt, respectively. }
\label{tab:notation}
\centering\footnotesize
{\setlength{\tabcolsep}{4pt}\renewcommand{\arraystretch}{1.12}
\begin{tabularx}{\linewidth}{@{}l>{\raggedright\arraybackslash}X@{\hspace{1.5em}}l>{\raggedright\arraybackslash}X@{}}
\toprule
Symbol & Meaning & Symbol & Meaning \\
\midrule
$\mathcal{W}$; $\mathcal{X}=\{x_1,\dots,x_N\}$ & world; its $N$ tasks & $\mathcal{M}$ & adaptation mechanism \\
$x^{(j)}$ & task at stream position $j$ & $\theta$ & current policy weights \\
$m$; $k$ & rounds per task; attempts per round & $c$; $h$ & retained context; teacher hindsight \\
$y_i\sim\pi_\theta(\cdot\mid c,x^{(j)})$ & multi-turn attempt $i$ & $p_j=\frac{1}{k}\sum_{i=1}^{k}s_{j,i}$ & first-round success rate on task $x^{(j)}$ \\
\addlinespace[3pt]
$o_i=(r_i,s_i,f_i)$ & reward, success indicator, feedback & $S=\frac{1}{N}\sum_{j=1}^{N}p_j$ & stream success rate \\
\addlinespace[3pt]
$E=\{(c,x^{(j)},y_i,o_i)\}_{i=1}^{k}$ & a round's experience & $\Delta S=S-S_{\mathrm{base}}$ & gain over the stateless baseline \\
\bottomrule
\end{tabularx}}
\end{table}

\subsection{Adaptation mechanisms}
\label{sec:interface}

We compare the stateless baseline with fourteen adaptation mechanisms, built from
four ICL policies and two weight-update rules. \textbf{ICL} replays prior attempts as
conversations, including environment observations and outcome feedback. Under a
bounded token allowance, two retention policies determine how context is managed:
\emph{\policy{FIFO}} evicts the oldest episodes to admit new ones, whereas
\emph{\policy{Prefix}} stops admitting episodes after the first overflow.
Two admission policies determine which attempts are eligible:
\emph{\policy{All}} selects the highest-reward attempt from every round, even if
it failed; \emph{\policy{Successful}} selects at most one successful attempt per
task, from the first round that produces one.

\looseness=-1 \textbf{GRPO} \citep{shao_deepseekmath_2024} updates a persistent LoRA
adapter \citep{hu_lora_2022} from each round's attempts: advantages are rewards relative to the
round's mean, and the loss is PPO's clipped surrogate \citep{schulman_ppo_2017}, without a KL
penalty or standard-deviation normalization.
\textbf{SDPO} \citep{hubotter_reinforcement_2026} trains a persistent LoRA adapter on
the agent's own attempts by minimizing the reverse KL divergence to a self-teacher's next-token
distributions, with no policy-gradient term. The self-teacher receives the agent's original context augmented with a
successful sibling attempt or, when none is available, the attempt's feedback.
Its weights start as the student's and then track them as an exponential moving
average. Each \textbf{hybrid} combines one ICL policy with one weight-update rule. Each round's
attempts feed both channels: the update trains the adapter on them, conditioned on the context
they were generated with, and the ICL policy chooses which of them to keep in the context for
later attempts.
Appendix~\ref{app:extended-framework} gives the formal interfaces and losses.

\subsection{Scoring}
\label{sec:protocol}

The first round's $k$ attempts at each task share the same context and weights, and
we score them \emph{before} their experience enters context or updates the weights.
Further rounds on that task supply adaptation experience but do not enter the main
score, so each task tests transfer from earlier tasks. Every mechanism uses the same task order, per-attempt limits, sampling
temperature, and source of feedback, learns only from its own attempts and the
environment's outcomes, and completes its updates before the next round.
Algorithm~\ref{alg:online-adaptation} gives the full loop. This score-then-update protocol follows the test-then-train scheme of
prequential evaluation \citep{dawid_prequential_1984,gama_evaluating_2013}: each task is first a test and then training data.

\paragraph{Metrics.}
\label{sec:metrics}
For task $j$, let $s_{j,i}$ indicate success of first-round attempt $i$, and let
$p_j=k^{-1}\sum_{i=1}^{k}s_{j,i}$. Across the world's $N$ tasks, we report
\begin{equation}
 S=\frac{1}{N}\sum_{j=1}^{N}p_j.
 \label{eq:metrics}
\end{equation}
$S$ denotes the success rate, averaged over tasks.
We compare mechanisms on the same stream, with $\Delta S$ denoting the
difference from the stateless baseline. Formal details are in Appendix~\ref{app:extended-framework};
implementation specifics are in Section~\ref{app:implementation}.

\algtext*{EndFor}
\algtext*{EndIf}
\algrenewcommand\algorithmicrequire{\textbf{Input:}}
\begin{algorithm}[!htbp]
\caption{Test-time adaptation on one world. }
\label{alg:online-adaptation}
\begin{algorithmic}[1]
\Require tasks $\mathcal{X}$, seed, mechanism $\mathcal{M}$, weights $\theta$, rounds per task $m$, attempts per round $k$
\State $(x^{(1)},\ldots,x^{(N)})\gets\textsc{Stream}(\mathcal{X},\text{seed})$
\For{$j=1,\ldots,N$} \Comment{each task once, in stream order}
  \For{$\ell=1,\ldots,m$} \Comment{rounds on the same task}
    \State $c\gets\mathcal{M}.\textsc{Prepare}(x^{(j)})$ \Comment{context from retained experience}
    \State $E\gets\{(c,x^{(j)},y_i,o_i)\}_{i=1}^{k}$ with $y_i\sim\pi_\theta(\cdot\mid c,x^{(j)})$ \Comment{$k$ multi-turn attempts}
    \If{$\ell=1$} \Comment{score before adapting}
      \State $p_j\gets\frac{1}{k}\sum_{i=1}^{k}s_{j,i}$
    \EndIf
    \State $\mathcal{M}.\textsc{Observe}(E)$ \Comment{update non-parametric state}
    \State $h\gets\mathcal{M}.\textsc{TeacherContexts}(E)$ \Comment{hindsight for SDPO's self-teacher}
    \State $\theta\gets\textsc{Update}(\theta,E,h)$ \Comment{GRPO or SDPO; unchanged for frozen weights}
  \EndFor
\EndFor
\State \Return $S=\frac{1}{N}\sum_j p_j$
\end{algorithmic}
\end{algorithm}

\section{Experiments}
\label{sec:experiments}

\subsection{Setup}

We evaluate Qwen3.5-9B on the four worlds of Section~\ref{sec:setting}.
We compare fourteen adaptation mechanisms and a frozen baseline, crossing five
context settings (no ICL and the four ICL policies) with frozen weights, SDPO, or
GRPO. All are evaluated using five stream seeds (0--4), with the same task order
for a given world and seed. We sample $k=16$ parallel attempts per round at temperature 1. Weight updates train
a rank-32 LoRA adapter with AdamW at learning rate $10^{-5}$.
We allocate separate token allowances to retained experience and each
new attempt, so filling the
retained-context allowance does not consume the new attempt's allowance.
Tables~\ref{tab:configs} and~\ref{tab:weight-update-settings} in Appendix~\ref{app:configs} give the full settings.

Following Section~\ref{sec:protocol}, we score only the first round of
each task, before adapting to that task, and average success over tasks within
each world. The overall mean weights the four worlds equally within each seed.

\subsection{Results}
\label{sec:results}

\begin{figure}[!htb]
\centering
\includegraphics[width=\linewidth]{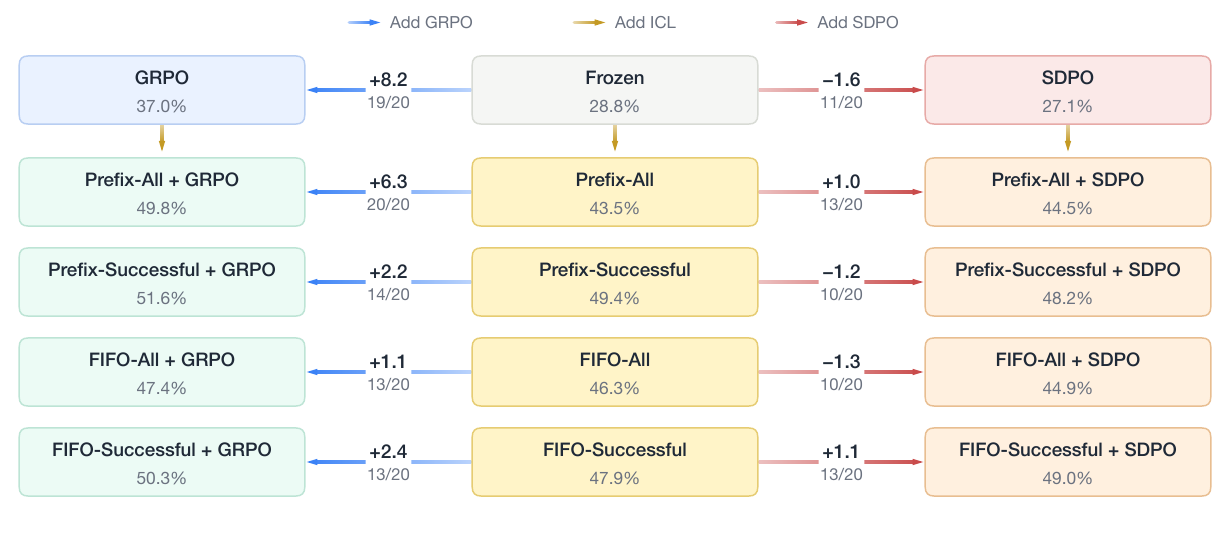}
\caption{\textbf{Mean gains and wins from adding GRPO or SDPO.}
Boxes show mean success rates. Yellow arrows indicate adding ICL with the
policies listed below; blue and red arrows add GRPO and SDPO, respectively.
Horizontal arrow labels show the mean success-rate gain in percentage points
(top) and wins out of 20 world--seed stream pairs (bottom), weighting all
streams equally.}
\label{fig:stream-win-rates}
\end{figure}

Table~\ref{tab:main} reports success rates for all mechanisms and the baseline,
with means and standard errors across the five seeds.
Figure~\ref{fig:stream-win-rates} complements these averages with paired
comparisons of mechanisms on the same ordered tasks within each world and seed.
Each arrow reports the mean gain and the number
of pairs with strictly higher success across the 20 world--seed pairs, showing both the size and consistency of
the gains. The figure compares adding GRPO or SDPO to Frozen and each ICL
policy. Complete comparisons, including gains from adding ICL, appear in
Appendix Figures~\ref{fig:stream-win-rates-grpo-full}
and~\ref{fig:stream-win-rates-sdpo}.

\begin{table}[!htbp]
\caption{\textbf{Success rates for the adaptation mechanisms and baseline.}
Mean success rates (\%) across five stream seeds, with standard errors in superscripts.
Rows are grouped as no adaptation, weight-only, ICL-only, and hybrid.
ICL policies are shown as retention\,\textperiodcentered\,admission.
No ICL with frozen weights is the baseline.}
\label{tab:main}
\label{tab:configuration-summary}
\centering\small
{\setlength{\tabcolsep}{2.5pt}\definecolor{segrey}{HTML}{6B7280}
\newcommand{\mse}[2]{\makebox[2.05em][r]{#1}\makebox[1.85em][l]{\textsuperscript{\color{segrey}$\pm$#2}}}
\begin{tabularx}{\linewidth}{@{}ll*{5}{>{\centering\arraybackslash}X}@{}}
\toprule
Weight update & ICL policy & \shortstack{SWE-bench\\Verified\\{\scriptsize (Xarray)}} & \shortstack{AppWorld\\{\scriptsize (File System)}} & \shortstack{ScienceWorld\\{\scriptsize (Measurement)}} & \shortstack{BIRD-Critic-\\SQLite\\{\scriptsize (Financial)}} & Mean \\
\midrule
\rowcolor{mechanismFrozen}Frozen & \textsf{No ICL} & \mse{13.4}{1.0} & \mse{57.7}{0.8} & \mse{15.6}{0.7} & \mse{28.5}{0.7} & \mse{28.8}{0.4} \\
\midrule
\rowcolor{mechanismSDPO}SDPO & \textsf{No ICL} & \mse{16.8}{0.8} & \mse{44.8}{1.2} & \mse{19.2}{0.6} & \mse{27.8}{0.6} & \mse{27.1}{0.3} \\
\rowcolor{mechanismGRPO}GRPO & \textsf{No ICL} & \mse{21.8}{1.3} & \mse{61.3}{1.3} & \mse{31.8}{1.6} & \mse{33.1}{1.1} & \mse{37.0}{0.7} \\
\midrule
\rowcolor{mechanismICL}Frozen & \policy{FIFO\,\textperiodcentered\,All} & \mse{39.4}{1.8} & \mse{66.9}{3.3} & \mse{50.3}{2.4} & \mse{28.5}{0.7} & \mse{46.3}{1.1} \\
\rowcolor{mechanismICL} & \policy{FIFO\,\textperiodcentered\,Successful} & \mse{35.1}{0.9} & \mse{77.9}{3.0} & \mse{46.1}{5.9} & \mse{32.6}{0.9} & \mse{47.9}{2.1} \\
\rowcolor{mechanismICL} & \policy{Prefix\,\textperiodcentered\,All} & \mse{35.6}{1.9} & \mse{62.6}{5.0} & \mse{48.0}{3.6} & \mse{27.9}{0.7} & \mse{43.5}{1.9} \\
\rowcolor{mechanismICL} & \policy{Prefix\,\textperiodcentered\,Successful} & \mse{33.4}{0.6} & \mse{\textbf{78.2}}{3.4} & \mse{53.2}{1.8} & \mse{32.7}{1.5} & \mse{49.4}{1.0} \\
\midrule
\rowcolor{mechanismICLSDPO}SDPO & \policy{FIFO\,\textperiodcentered\,All} & \mse{38.4}{1.1} & \mse{53.7}{4.3} & \mse{56.6}{1.8} & \mse{31.1}{1.2} & \mse{44.9}{0.8} \\
\rowcolor{mechanismICLSDPO} & \policy{FIFO\,\textperiodcentered\,Successful} & \mse{36.2}{1.6} & \mse{75.2}{3.5} & \mse{52.8}{6.4} & \mse{31.8}{1.2} & \mse{49.0}{1.4} \\
\rowcolor{mechanismICLSDPO} & \policy{Prefix\,\textperiodcentered\,All} & \mse{33.7}{4.0} & \mse{59.0}{3.7} & \mse{55.5}{5.1} & \mse{29.8}{1.2} & \mse{44.5}{1.0} \\
\rowcolor{mechanismICLSDPO} & \policy{Prefix\,\textperiodcentered\,Successful} & \mse{35.2}{1.4} & \mse{76.8}{1.5} & \mse{50.2}{4.3} & \mse{30.6}{0.8} & \mse{48.2}{1.6} \\
\cmidrule(lr){1-7}
\rowcolor{mechanismICLGRPO}GRPO & \policy{FIFO\,\textperiodcentered\,All} & \mse{38.9}{1.8} & \mse{66.1}{1.5} & \mse{50.1}{2.5} & \mse{34.4}{2.1} & \mse{47.4}{0.6} \\
\rowcolor{mechanismICLGRPO} & \policy{FIFO\,\textperiodcentered\,Successful} & \mse{36.6}{1.8} & \mse{76.4}{1.2} & \mse{55.2}{2.4} & \mse{33.0}{0.4} & \mse{50.3}{0.6} \\
\rowcolor{mechanismICLGRPO} & \policy{Prefix\,\textperiodcentered\,All} & \mse{\textbf{39.9}}{1.9} & \mse{69.4}{3.5} & \mse{55.9}{2.6} & \mse{34.0}{1.6} & \mse{49.8}{1.0} \\
\rowcolor{mechanismICLGRPO} & \policy{Prefix\,\textperiodcentered\,Successful} & \mse{37.6}{1.4} & \mse{74.5}{4.1} & \mse{\textbf{58.1}}{1.2} & \mse{\textbf{36.3}}{1.6} & \mse{\textbf{51.6}}{1.3} \\
\bottomrule
\end{tabularx}
}
\end{table}

\textbf{ICL provides the largest standalone gain.}
With frozen weights, the four ICL policies raise success from the baseline's
28.8\% to 43.5--49.4\%.
\policy{Prefix}--\policy{Successful} is the strongest ICL policy.
Both \policy{Successful}-admission policies beat the baseline in all 20
stream pairs (Appendix Figure~\ref{fig:stream-win-rates-grpo-full}), whereas \policy{All}
admission occasionally lowers success on AppWorld or BIRD.
GRPO alone improves on the baseline in 19 of 20 pairs, with a smaller
overall gain than any ICL policy, while SDPO alone has a lower mean.
ICL has the higher standalone mean in three worlds; on BIRD, which
has roughly twice as many tasks as the other worlds, GRPO has a slightly higher
mean over the full stream.

\textbf{Adding GRPO improves the overall mean under every ICL policy; SDPO has mixed effects.}
Adding GRPO to ICL raises the overall mean under all four ICL policies,
most for \policy{Prefix}--\policy{All}: a gain of 6.3 percentage
points, with improvements in all 20 stream pairs. Under the other three policies, GRPO
improves success in 13--14 of 20 pairs (Figure~\ref{fig:stream-win-rates}).
Adding SDPO instead raises the mean under two ICL policies and lowers it under the other two
(Figure~\ref{fig:stream-win-rates}), so using both channels does not
by itself ensure an improvement over ICL alone. Nevertheless, the three highest
mean success rates all come from ICL with GRPO, led by
ICL(\policy{Prefix}--\policy{Successful}) $+$ GRPO at 51.6\%.
Additional seed-4 experiments on two larger worlds, scikit-learn and
Toxicology, also place ICL$+$GRPO highest among the six tested mechanisms
(Appendix~\ref{app:additional-worlds}).

\FloatBarrier
\section{Individual Contributions of Context and Weights}
\label{sec:single-channel}
\label{sec:icl-effects}
\label{sec:weight-effects}

What does each adaptation channel contribute to test-time adaptation?
Beyond the success rates in Section~\ref{sec:experiments}, we examine how context
and weight updates separately change the agent's outcomes, interaction length,
and behavior relative to the frozen baseline. For this analysis, we use the ICL policy with the highest mean success rate
in each benchmark (Table~\ref{tab:main}): \policy{FIFO}--\policy{All} on SWE-bench and
\policy{Prefix}--\policy{Successful} elsewhere. Appendix
Figure~\ref{fig:icl-cumulative-gain-context} compares cumulative gains and
retained context for all four ICL policies.

\subsection{Outcomes and Interaction Length}
\label{sec:single-channel-outcomes}

\begin{figure}[!htb]
\centering
\includegraphics[width=\linewidth]{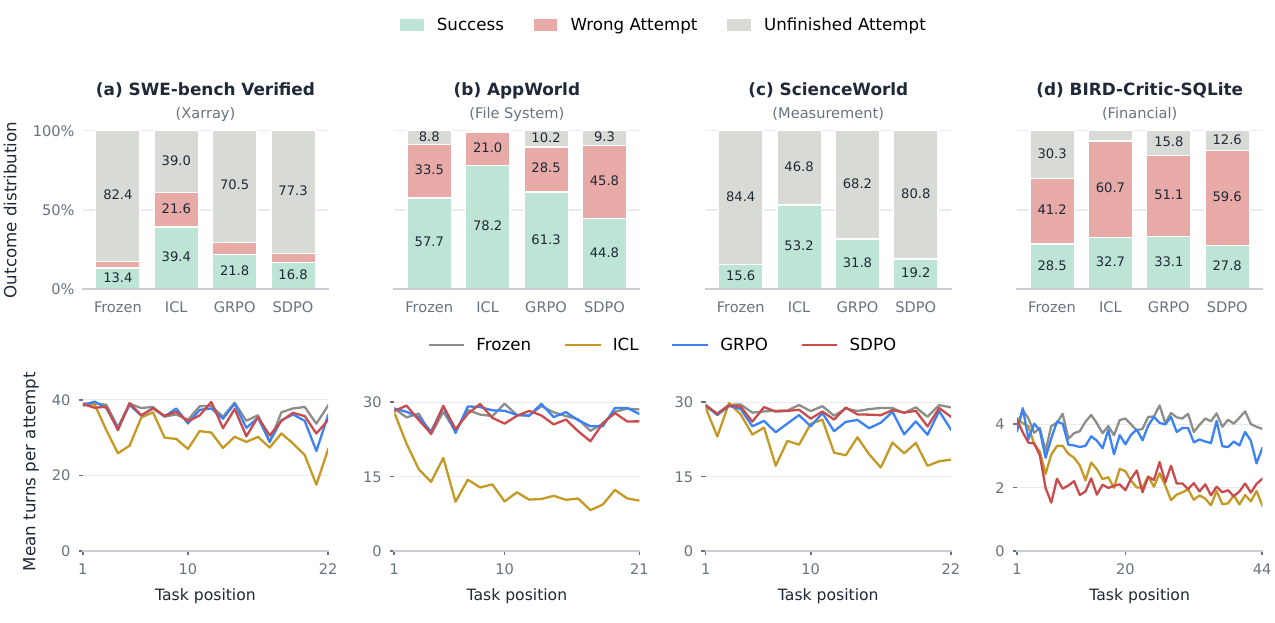}
\caption{\textbf{How single-channel adaptation changes outcomes and attempt length.}
Top: mean proportions of successful, wrong, and unfinished attempts. An unfinished attempt ends without
success or submission. ScienceWorld ends on reaching the goal and has no
separate submission decision. Bottom: mean turns per attempt at each task
position. Results weight the
five stream seeds equally, with 16 attempts per task and stream.
}
\label{fig:single-channel-outcomes-turns}
\end{figure}

\textbf{ICL improves both completion and success.}
The outcome distributions in Figure~\ref{fig:single-channel-outcomes-turns}
(top) show that ICL substantially reduces unfinished attempts in every
benchmark, nearly eliminating them on AppWorld. On SWE-bench and BIRD, both
successful and wrong submissions become more common.
AppWorld instead improves both completion and correctness, with fewer wrong
submissions. GRPO raises success on all four benchmarks, with smaller
completion changes. SDPO's effects are mixed: it increases wrong submissions
on AppWorld and reduces unfinished attempts on BIRD without improving success.

\textbf{ICL shortens interaction more consistently than weight updates.}
Figure~\ref{fig:single-channel-outcomes-turns} (bottom) shows a substantial
decline in ICL's mean turns along every stream, most pronounced on AppWorld
and BIRD. Successful ICL attempts are also shorter than successful baseline
attempts in every benchmark (Appendix
Figure~\ref{fig:single-channel-conditional-turns}), so the lower overall
mean is not solely due to fewer unfinished attempts. SDPO also sharply shortens BIRD attempts despite its lack of a success-rate gain.

\subsection{Behavioral Changes}
\label{sec:single-channel-behavior}

\begin{figure}[!htb]
\centering
\includegraphics[width=\linewidth]{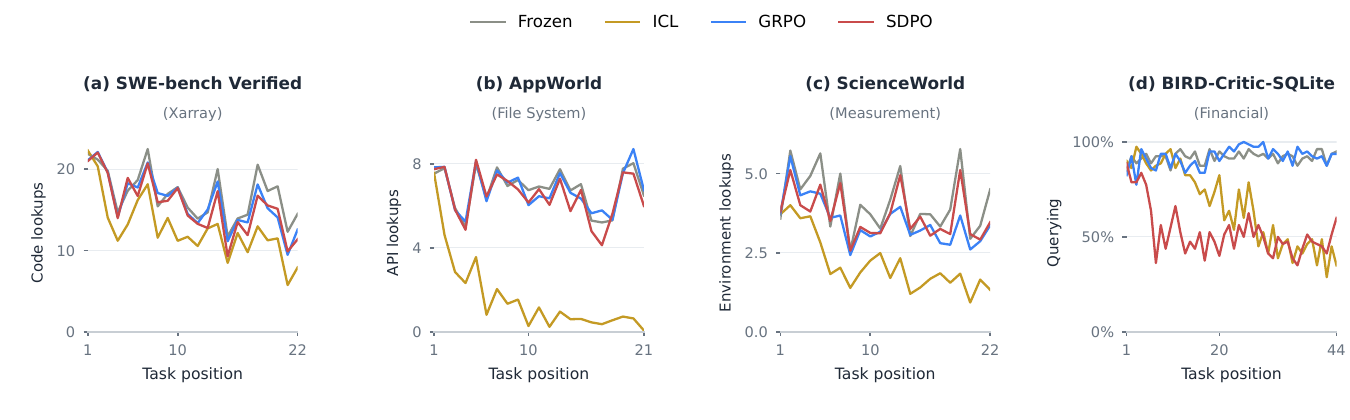}
\caption{\textbf{How single-channel adaptation changes behavior along the stream.}
SWE-bench: mean code-lookup turns per attempt. AppWorld: mean turns containing API-documentation lookups. ScienceWorld: mean explicit \texttt{look around} actions per attempt (environment lookups). BIRD: percentage of all attempts with at least one executed exploratory SQL
query. All panels weight the five
seeds equally at each position. }
\label{fig:single-channel-behavior}
\end{figure}

\begin{figure}[p]
\centering
\includegraphics[width=\linewidth,height=0.83\textheight,keepaspectratio]{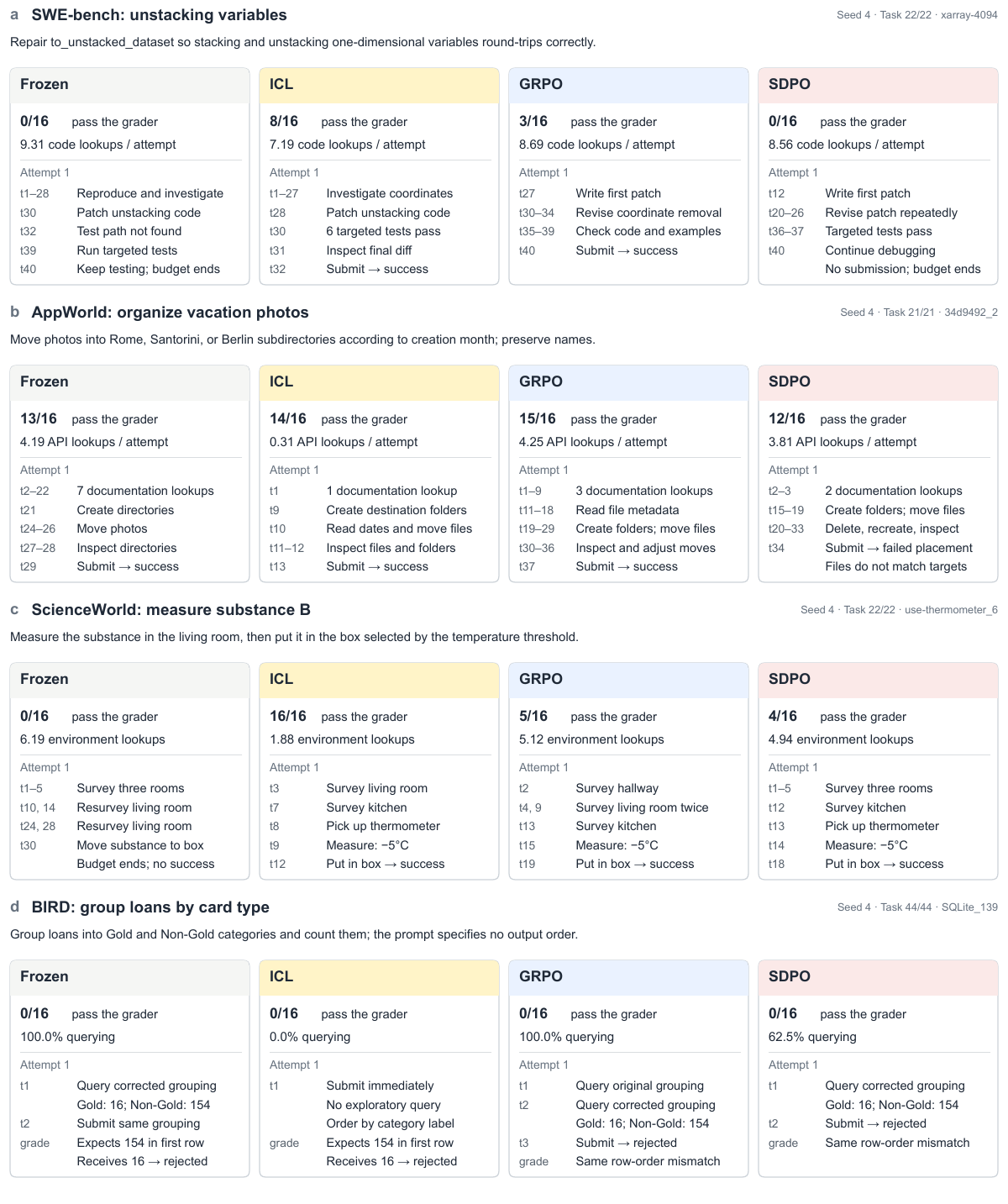}
\caption{\textbf{Final-task examples across all four benchmarks in seed~4.}
Each column summarizes all 16 attempts: graded successes and
mean code-lookup turns (SWE-bench), mean API lookups (AppWorld),
mean environment lookups (ScienceWorld), or the percentage of all
attempts with at least one executed exploratory SQL query (BIRD).
Below the divider, excerpts summarize attempt~1; $t$ denotes the turn. }
\label{fig:single-channel-traces}
\end{figure}

We next examine how the agent uses its turns, looking more closely at
which interactions change alongside the outcomes and attempt lengths.
Inspection of execution traces motivated the four behavioral measures in
Figure~\ref{fig:single-channel-behavior}: code lookups, API lookups,
environment lookups, and database querying. These measures track how each
adaptation channel changes the agent's information gathering along the
stream. Figure~\ref{fig:single-channel-traces} provides concrete examples
using the final task of seed~4 in each benchmark.

\begin{samepage}
\textbf{SWE-bench: ICL reduces code lookup.}
We count each turn that searches or displays repository files, excluding
test runs, diffs, and reads used only for editing, and report the mean count
per attempt.
ICL uses fewer turns to search for or read repository code, whereas the
weight-only mechanisms remain closer to the baseline
(Figure~\ref{fig:single-channel-behavior}a). The reduction holds
in every seed and is consistent with less repository exploration within
ICL's shorter attempts. The unstacking example
(Figure~\ref{fig:single-channel-traces}a) shows the same reduction
across attempts.
\par
\end{samepage}

\textbf{AppWorld: retained experience reduces repeated API lookup.}
We count turns containing API-documentation lookups and
report the mean per attempt. Under ICL, these lookups rapidly decline toward
zero, alongside shorter attempts and higher success. GRPO and SDPO remain
close to the baseline (Figure~\ref{fig:single-channel-behavior}b).
This is consistent with ICL reusing API
procedures from retained experience instead of rediscovering them for each
task. In the vacation-photo example
(Figure~\ref{fig:single-channel-traces}b), the displayed ICL attempt uses only
one documentation lookup and combines metadata reads and file moves into a
compact sequence.

\Needspace{4\baselineskip}
\textbf{ScienceWorld: ICL reduces environment lookups.}
We count explicit \texttt{look around} actions per attempt, which reveal
surrounding objects and exits. ICL reduces these surveys substantially,
while GRPO and SDPO remain closer to Frozen
(Figure~\ref{fig:single-channel-behavior}c).
In the displayed thermometer attempt (Figure~\ref{fig:single-channel-traces}c),
ICL surveys only the living room and kitchen before measuring the substance
and placing it in the correct box. Frozen repeatedly surveys the living room;
GRPO and SDPO each use four surveys, compared with ICL's two.

\textbf{BIRD: shorter attempts can omit database checks.}
We measure the fraction of attempts containing at least
one executed exploratory SQL query.
ICL increasingly submits without an exploratory query, and SDPO also
reduces querying. GRPO maintains querying in a larger
fraction of attempts than ICL or SDPO (Figure~\ref{fig:single-channel-behavior}d).
Thus, unlike fewer documentation
lookups on AppWorld, shorter BIRD attempts also mean fewer opportunities to
test the proposed answer. The final-task example
(Figure~\ref{fig:single-channel-traces}d) makes the distinction explicit:
ICL submits immediately in every attempt, while Frozen and GRPO always query
first.

\Needspace{5\baselineskip}
\textbf{ICL can reuse environment knowledge to act more efficiently.}
Retained experiences give ICL access to information about the environment's latent
structure, reducing the need to rediscover it on each task. This is consistent
with its fewer code, API, and environment lookups
(Figure~\ref{fig:single-channel-behavior}a--c),
shorter attempts and higher completion and success
(Figure~\ref{fig:single-channel-outcomes-turns}). Even successful attempts
become shorter (Appendix Figure~\ref{fig:single-channel-conditional-turns}).
Weight updates offer a useful contrast.
Both methods learn from the agent's own attempts. GRPO's
loss uses the probabilities of sampled tokens, whereas SDPO also matches
self-teacher probabilities for non-sampled tokens. GRPO reinforces higher-reward traces, which still include exploration, consistent with its lookup and querying metrics
remaining closer to Frozen than ICL even as success improves
(Figure~\ref{fig:single-channel-behavior}). However, SDPO's self-teacher uses extra information in context
and could pull the student toward token choices that skip exploration, consistent
with its ICL-like reduction in querying on BIRD
(Figure~\ref{fig:single-channel-behavior}d).

\textbf{Greater efficiency can come at the cost of verification.}
ICL can reuse experience to avoid rediscovering the environment's shared
structure. However, finding
information relevant to the current task and checking a proposed solution may
still require new interactions. ICL risks extending the shortcut of skipping
familiar exploration to these task-specific steps. BIRD illustrates this risk:
ICL increasingly submits SQL without first querying the database
(Figures~\ref{fig:single-channel-behavior}d and~\ref{fig:single-channel-traces}d),
skipping opportunities to inspect relevant data or test its proposed answer.
ICL improves success but also produces more wrong
submissions
(Figure~\ref{fig:single-channel-outcomes-turns}d, top). Effective adaptation
should therefore reuse knowledge of the environment while preserving the
exploration and verification needed for each new task.

\FloatBarrier
\section{When Context and Weight Updates Complement Each Other}
\label{sec:hybrid-mechanisms}
\label{sec:memory-policy}
\label{sec:analysis} %

When does combining context and weight adaptation help, and how do the two
channels complement each other? Building on the single-channel analysis in
Section~\ref{sec:single-channel}, we examine how adding weight updates to ICL
changes outcomes and behavior across benchmarks. Adding GRPO improves the
overall mean under every ICL policy (Table~\ref{tab:main};
Figure~\ref{fig:stream-win-rates}). We focus on \policy{Prefix}--\policy{All},
where this gain is largest and positive in all 20 benchmark--seed streams,
and compare it with its GRPO and SDPO hybrids, alongside the frozen, no-ICL
baseline.

\label{sec:channel-interaction}
\label{sec:behavior}

\begin{figure}[!htb]
\centering
\includegraphics[width=\linewidth]{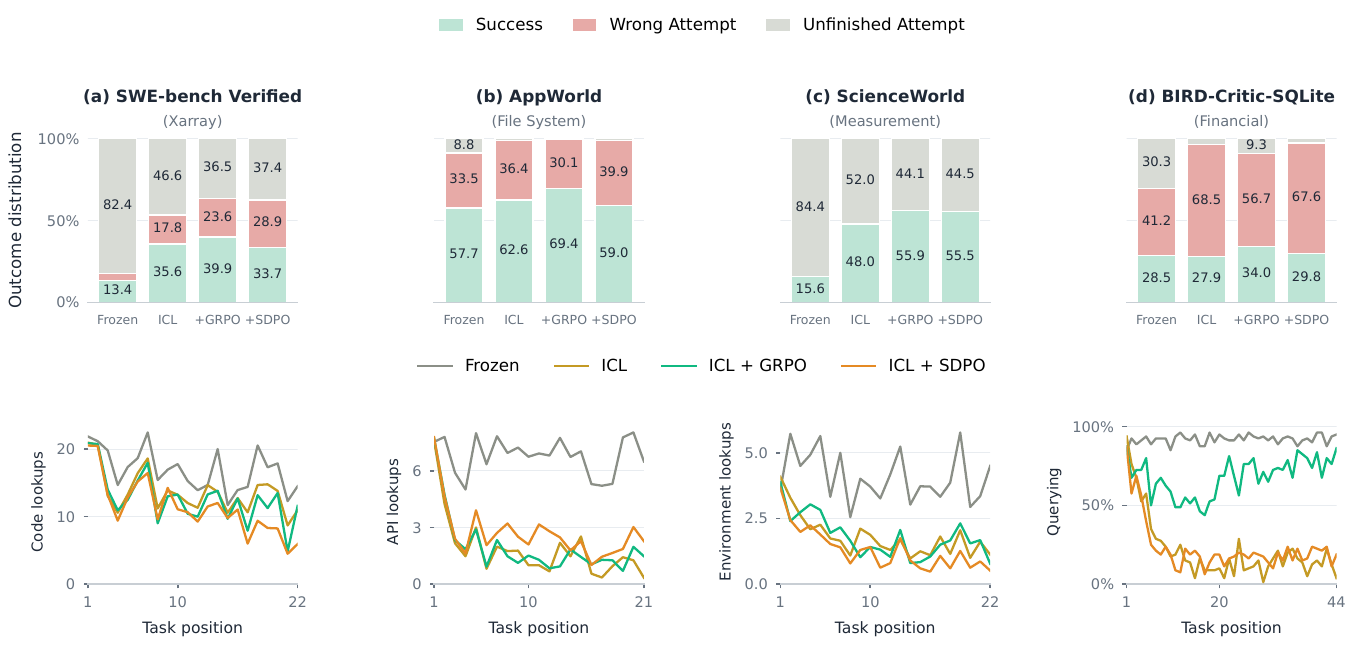}
\caption{\textbf{Hybrid outcomes and exploration: GRPO preserves reduced lookups while restoring querying.}
Top: mean proportions of successful, wrong, and unfinished attempts.
An unfinished attempt ends without success or submission.
Bottom: mean code-lookup turns per attempt on SWE-bench, mean turns containing
API-documentation lookups on AppWorld, mean explicit \texttt{look around}
actions on ScienceWorld, and percentage of all attempts with at least one
executed exploratory SQL query on BIRD.
All panels weight the five seeds equally.}
\label{fig:prefix-all-outcomes-turns}
\label{fig:prefix-all-exploration}
\end{figure}

\Needspace{5\baselineskip}
\textbf{Weight updates change the outcome distribution differently across benchmarks.}
On SWE-bench, adding either GRPO or SDPO reduces unfinished attempts, but
success increases only with GRPO; under SDPO, fewer unfinished attempts
accompany more wrong submissions and lower success
(Figure~\ref{fig:prefix-all-outcomes-turns}, top). On AppWorld, GRPO reduces
wrong submissions, whereas SDPO increases them. On ScienceWorld, both yield
similar success gains. On BIRD, GRPO substantially reduces wrong submissions,
with both successful and unfinished attempts increasing; SDPO produces a
smaller shift toward success. We next examine which interactions change
alongside these outcomes.

\textbf{Adding GRPO preserves ICL's reduced exploration.}
Revisiting the behavioral metrics from Section~\ref{sec:single-channel},
we find that adding GRPO retains ICL's reduced code, API, and environment
lookups on SWE-bench, AppWorld, and ScienceWorld
(Figure~\ref{fig:prefix-all-exploration}a--c, bottom). Success therefore
improves while these lookup rates remain low, consistent with continued
reuse of environment knowledge from retained experience.

\begin{figure}[!htb]
\centering
\includegraphics[width=\linewidth]{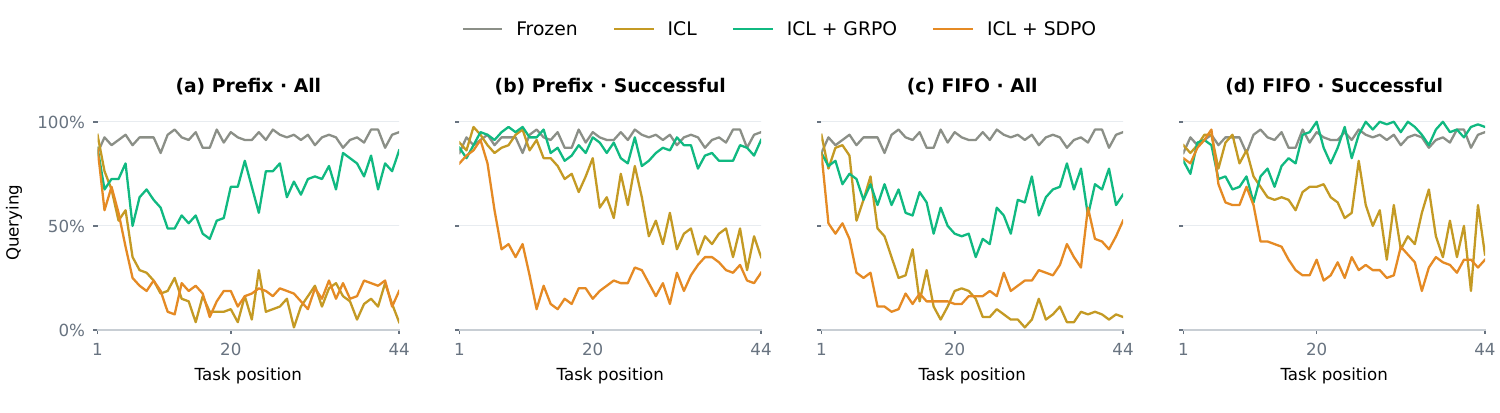}
\caption{\textbf{GRPO maintains higher late-stream querying across ICL policies on BIRD.}
Each panel compares ICL with its GRPO and SDPO hybrids under one ICL policy;
Frozen is repeated for reference. Querying counts all attempts with at least
one executed exploratory SQL query. Rates average 16 attempts per task in a stream.}
\label{fig:bird-querying-all-policies}
\end{figure}

\textbf{BIRD: adding GRPO restores querying.}
The clearest departure from ICL appears in BIRD's querying metric
(Figure~\ref{fig:prefix-all-exploration}d, bottom). With ICL$+$GRPO, querying
initially declines as it does with ICL, but then recovers toward the frozen
baseline as the stream progresses. Higher late-stream querying with GRPO also appears under all four
ICL policies (Figure~\ref{fig:bird-querying-all-policies}). Unlike looking up familiar
repository structure or API documentation, executing a query can provide
feedback about the current task and proposed answer. In the displayed
final-task attempt, ICL submits immediately, whereas ICL$+$GRPO executes
its query before submission (Figure~\ref{fig:hybrid-traces-main}d). This
distinction motivates examining whether adding GRPO also increases
task-specific inspection and verification in the other benchmarks.

\begin{figure}[b]
\centering
\includegraphics[width=\linewidth]{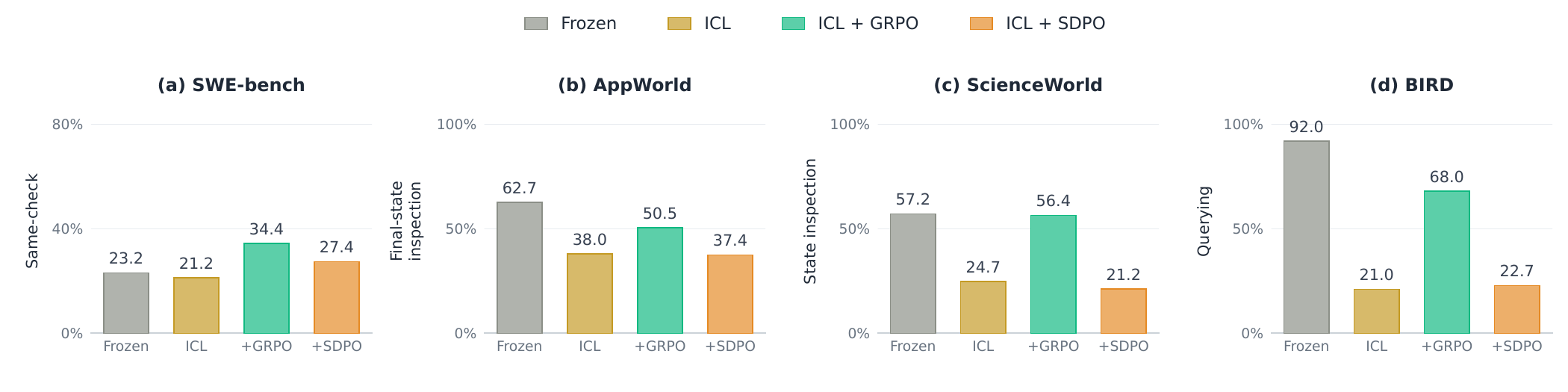}
\caption{\textbf{How weight updates change ICL's task-specific inspection and verification.}
(a) Rechecking after edits on SWE-bench; (b) final-state inspection on
AppWorld; (c) inspection during melting-point experiments on ScienceWorld;
(d) querying on BIRD. Bars show rates (\%) averaged equally over five seeds.}
\label{fig:prefix-all-behavior}
\end{figure}

\begin{figure}[p]
\centering
\includegraphics[width=\linewidth,height=0.83\textheight,keepaspectratio]{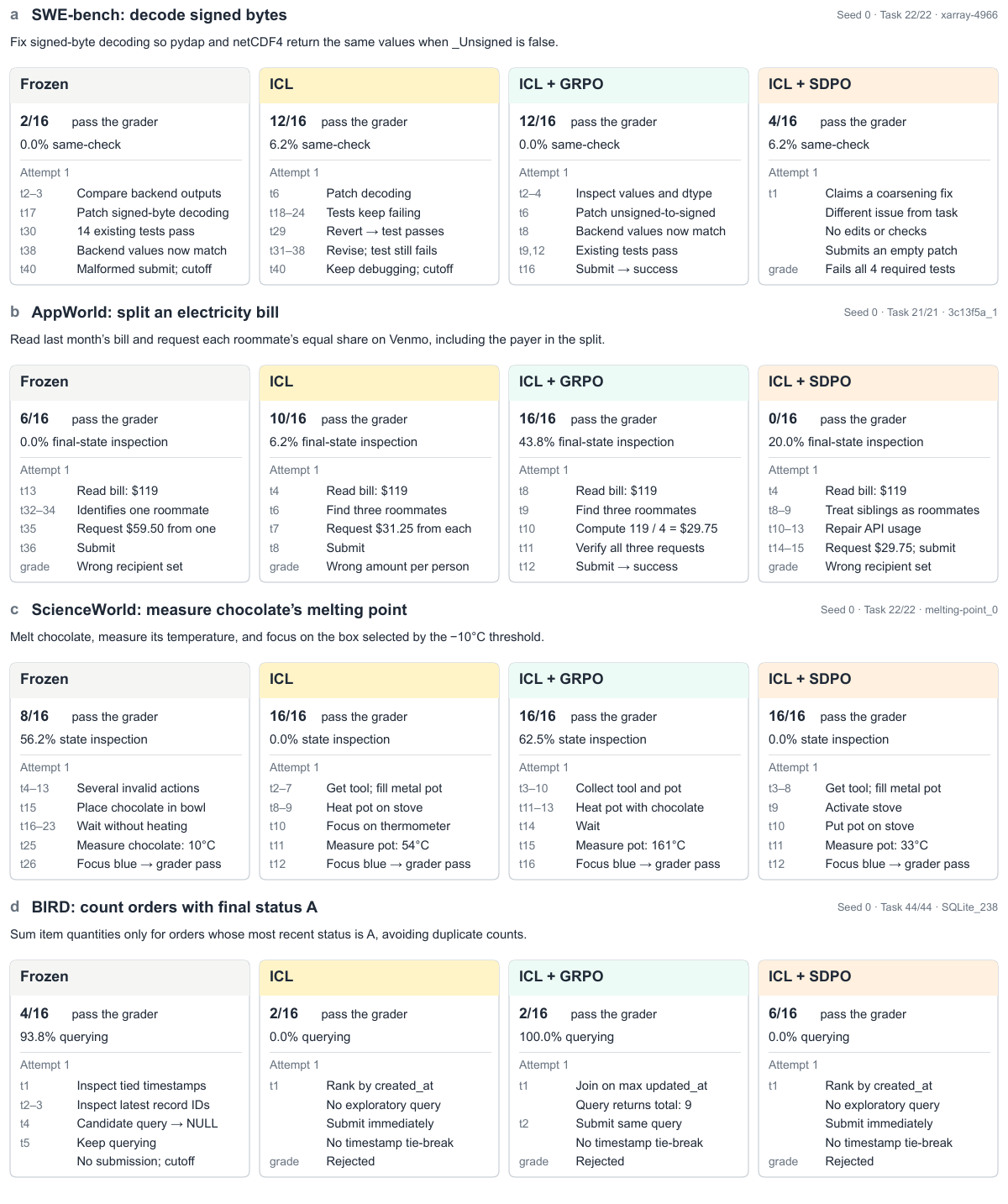}
\caption{\textbf{How weight updates change ICL behavior: final-task examples across all four benchmarks in seed~0.}
Each column summarizes all 16 attempts: graded successes and
same-check rates (SWE-bench), final-state inspection (AppWorld),
state inspection (ScienceWorld), or querying (BIRD), using the metrics of
Figure~\ref{fig:prefix-all-behavior}.
Below the divider, excerpts summarize attempt~1; $t$ denotes the turn. }
\label{fig:hybrid-traces-main}
\label{fig:prefix-all-traces}
\end{figure}

\Needspace{4\baselineskip}
\textbf{SWE-bench: fewer code lookups, more rechecking after edits.}
Knowing where to edit does not establish that a new patch works.
We measure the fraction of attempts that run the same Python probe or pytest
selection before and after an attempted source edit.
Both weight updates increase this rate relative to ICL, with a larger
increase under GRPO (Figure~\ref{fig:prefix-all-behavior}a).
Together with the reduced code lookups above, this distinguishes revisiting
repository information from rechecking behavior after changing the code.
The latter provides an opportunity to assess the current patch.

\textbf{AppWorld: fewer API lookups, more final-state inspection.}
An API procedure's effects depend on the current task and data. Among
attempts containing a state-changing API call, \emph{final-state inspection}
is the fraction that read task data after the last such call. Adding GRPO
increases this rate while preserving ICL's reduced API lookups; SDPO remains
close to ICL (Figure~\ref{fig:prefix-all-exploration}b, bottom, and
Figure~\ref{fig:prefix-all-behavior}b). In the bill-splitting example, ICL
submits immediately after requesting payments, whereas ICL$+$GRPO reads back
the requests before submitting (Figure~\ref{fig:hybrid-traces-main}b).

\Needspace{11\baselineskip}
\textbf{ScienceWorld: fewer environment lookups, more inspection during experiments.}
Among melting-point attempts, we measure the fraction with successful \texttt{look at} or
\texttt{look in} inspection after heating or cooling starts and before the
first box-directed action. Adding GRPO raises this rate toward Frozen while
preserving reduced \texttt{look around} use; SDPO shows no comparable increase
(Figure~\ref{fig:prefix-all-exploration}c, bottom, and Figure~\ref{fig:prefix-all-behavior}c).
Across the final task's 16 attempts, ICL$+$GRPO shows state inspection,
whereas ICL and ICL$+$SDPO show none (Figure~\ref{fig:hybrid-traces-main}c).

\FloatBarrier

\section{Discussion and Limitations}
\label{sec:discussion}

\textbf{How reinforcement learning and self-distillation interact with ICL.}
Retained traces provide environment knowledge and examples of how to act,
but a useful procedure or shortcut on one task may be insufficient on
another. GRPO applies reinforcement learning with verifiable rewards to the
policy conditioned on retained experience, potentially correcting behavior
that the context encourages but the current task's reward does not favor.
Our observations are consistent with this interpretation: adding GRPO
preserves reduced environment lookups while increasing task-specific
inspection and verification (Figures~\ref{fig:prefix-all-exploration}
and~\ref{fig:prefix-all-behavior}). Self-distillation also uses environmental
feedback, but its supervision depends on how a self-teacher uses additional
hindsight in context. This reliance on in-context learning could make
self-distillation susceptible to some of the same limitations as ICL's use
of retained experience, reinforcing similar shortcuts rather than correcting
them. This remains a hypothesis; studying when the self-teacher provides
useful corrections could clarify what limits self-distillation and how to
make it more effective alongside ICL.

\textbf{How context management shapes the benefits of weight updates.}
Context management determines both what experience the agent can reuse and
the context on which its weight updates are conditioned. An effective
retention policy for ICL alone may therefore differ from one that works
well with weight updates. \policy{Prefix} keeps retained history fixed after
the first context-budget overflow, potentially allowing updates to refine
behavior under a stable context. \policy{FIFO} continually introduces new
experience, but also changes the context under which previously learned
behavior is applied. If the benefits of weight updates survive eviction,
\policy{FIFO} could extend the usefulness of experience beyond the context
window; if those benefits depend strongly on retained episodes, a stable
context may be more effective. GRPO adds more to \policy{Prefix} than
\policy{FIFO} under \policy{All} admission, but the gains are similar under
\policy{Successful} admission (Figure~\ref{fig:stream-win-rates} and
Appendix Figure~\ref{fig:hybrid-grpo-context}), suggesting that admission and retention
should be studied jointly with weight updates. These comparisons do not
isolate the effect of context management because each run develops its own
trajectories and context. Evaluating the same updated weights under retained,
replaced, or removed experience could clarify how to manage context so that
the two channels work effectively together.

\textbf{Scope and limitations.}
We study one model on subsets of four benchmarks, with five task orders,
raw-trajectory context policies, and specific update recipes. Our metrics
need not capture all sources of improvement or generalize to every ICL
policy. These behavioral observations do not establish what causes the
success gains. Inspection and verification provide opportunities for
feedback, but our metrics do not establish whether the agent uses that
feedback correctly. Interventions that selectively require or suppress
these checks would help test their contribution. Longer streams, stronger
memory selection or compression, and other models may change the balance
between channels. Our main evaluation measures online improvement within a
deployment environment, not retention of general capabilities.
Appendix~\ref{app:transfer} separately examines transfer and component
ablations of final seed-4 agents.

\section{Conclusion}
\label{sec:conclusion}

We study how context and weight updates contribute to agent adaptation
after deployment, using a shared online protocol and an extensible evaluation
framework. Across four benchmarks, ICL provides the largest standalone gain,
while adding GRPO improves the overall mean under every ICL policy and yields
the strongest hybrid. Our behavioral analyses show that retained experience
reduces repeated exploration but can also encourage skipping checks needed
for the current task. In our detailed hybrid comparison, adding GRPO
preserves reduced lookups while increasing task-specific inspection and
verification; SDPO does not consistently reproduce this combination. These
findings suggest that reinforcement learning with verifiable rewards can
refine how agents use retained experience, motivating the joint study of
context management and weight updates to make the two channels work
effectively together.

\clearpage

\bibliography{references}
\bibliographystyle{plainnat}

\clearpage
\appendix

\section{Experimental Setup and Implementation Details}
\label{app:extended-framework}
\label{app:protocol}

This appendix specifies the adaptation mechanisms, benchmark settings, and
implementation choices for the main four-world online experiments. The online
protocol and scoring rule are given in Section~\ref{sec:setup} and
Algorithm~\ref{alg:online-adaptation}. Additional environments and evaluations
of final agents are described separately in Appendices~\ref{app:additional-worlds}
and~\ref{app:transfer}.

\subsection{Framework interfaces and adaptation mechanisms}
\label{app:mechanism-formalism}
\label{app:world-formalism}

A benchmark registers four contracts: a \emph{world} defines the tasks and shared
interaction rules; a \emph{stream} orders those tasks; a \emph{task} supplies its
prompt and grading specification; and a \emph{session} executes one isolated
attempt. A mechanism implements \textsc{Prepare} to return context messages,
\textsc{Observe} to receive completed attempts and outcomes and update its
non-parametric state, and, for self-distillation, \textsc{TeacherContexts} to
construct hindsight contexts. Save and restore hooks preserve its state across
checkpoints. The trainer supplies a budget check for context preparation and
teacher-context construction.

For a round on task $x^{(j)}$, the experience is
$E=\{(c,x^{(j)},y_i,o_i)\}_{i=1}^{k}$, where all $k$ attempts share the
context $c$ and policy weights used for generation. Weight updates minimize
\begin{equation}
\mathcal{L}(\theta;E)=
\lambda_{\mathrm{SDPO}}\mathcal{L}_{\mathrm{SDPO}}
\bigl(\theta;E,\textsc{TeacherContexts}(E)\bigr)+
\lambda_{\mathrm{GRPO}}\mathcal{L}_{\mathrm{GRPO}}(\theta;E).
\label{eq:update}
\end{equation}
Both coefficients are zero for the stateless baseline and ICL-only mechanisms. GRPO sets
$(\lambda_{\mathrm{SDPO}},\lambda_{\mathrm{GRPO}})=(0,1)$; SDPO sets them to $(1,0)$.
The LoRA adapter and optimizer state persist across tasks. In hybrids, updates
use attempts generated with retained experience in context, and the updated policy generates
later context candidates. All learning uses the agent's own attempts and outcomes.

\paragraph{ICL policies.}
Episodes contain actions, environment observations, and final outcome feedback,
replayed oldest first between the system prompt and the current task. \policy{All}
admission selects the highest-reward attempt from each round, including failures.
\policy{Successful} admission selects at most one successful attempt per task, from the
first round containing a success. It therefore changes both outcome filtering
and the number of episodes a task may contribute. \policy{FIFO} removes oldest whole
episodes until the context fits; \policy{Prefix} permanently stops admitting episodes at
the first overflow. Neither uses summarization, retrieval, or an external judge.
Reward ties, successful-attempt selection, and SDPO donor selection follow the
trainer's attempt order, which verl rearranges to balance sequence lengths.

\paragraph{GRPO.}
The policy-gradient objective uses each attempt's reward minus its round's mean
reward as the advantage, without standard-deviation normalization. The loss is
averaged over generated tokens and uses clipping without a KL penalty toward the
base model. Table~\ref{tab:weight-update-settings} gives clipping and optimization settings.

\paragraph{SDPO.}
The self-teacher starts as a copy of the student. After each step it is updated
as $\bar\theta\leftarrow(1-\beta)\bar\theta+\beta\theta$, with $\beta=0.01$;
the two LoRA factors are averaged separately and the teacher is checkpointed.
For each attempt, hindsight is a successful sibling from the same round,
excluding the attempt itself, or, if none exists and the attempt failed, its own
feedback. Attempts with neither contribute zero loss. The teacher sees the
student's original prompt, including retained context in hybrids, with hindsight added to
the task message. Hindsight is shortened with an explicit marker if necessary;
the original prompt is preserved.

The loss is reverse KL, $\mathrm{KL}(\pi_\theta\,\|\,\pi_{\bar\theta})$,
on the student's generated positions, evaluated over its top 20 tokens and one
bucket for the remaining probability mass. Each token's divergence is multiplied
by two detached importance weights, each capped at 2: the student-to-generating
policy probability ratio and verl's inference-to-trainer probability correction.
The loss averages over generated tokens within an attempt and then over all
$k$ attempts, including zero contributions. Support and teacher scores are
computed once per round and are not refreshed after an update, so SDPO takes one
optimizer step per round. This implementation uses no policy-gradient term.

\subsection{Benchmark environments, feedback, and task limits}
\label{app:worlds}

Each world is shuffled once per seed, and each task appears once in the stream.
Multiple rounds on that task supply adaptation experience; only the first round
contributes to reported success rate.

\begin{table}[!htbp]
\caption{\textbf{Benchmark settings and feedback.} Success is each benchmark's verdict (for ScienceWorld,
AgentBoard's subgoal annotations \citep{ma_agentboard_2024}); reward is the learning signal of GRPO and the
selection rule of the \policy{All}-admission ICL policies, whereas \policy{Successful} admission selects
by success.}
\label{tab:worlds}
\centering
\small
\begin{tabular}{@{}>{\raggedright\arraybackslash}p{2.2cm}>{\raggedright\arraybackslash}p{3.0cm}>{\raggedright\arraybackslash}p{2.5cm}c>{\raggedright\arraybackslash}p{3.4cm}@{}}
\toprule
World & Tasks & Action & Turns & Success, reward, feedback \\
\midrule
SWE-bench Verified: \world{Xarray} & 22 instances from \texttt{pydata/xarray} & one block of
  shell commands per turn, then a submission & 40 & issue resolved; binary reward;
  feedback reports how many of the required tests pass \\
AppWorld: \world{File System} & 21 train and dev tasks whose first required app is the file
  system (7 templates, 3 variants each) & Python code calling app APIs in a persistent
  interpreter & 40 & all task requirements pass; binary reward; feedback reports how
  many requirements passed and the first failure \\
ScienceWorld: \world{Measurement} & 22 AgentBoard instances of the thermometer and
  melting-point tasks & one text command & 30 & all annotated subgoals reached; reward
  is the fraction reached; feedback reports it \\
BIRD-Critic-SQLite: \world{Financial} & 44 tasks on the \texttt{financial} SQLite database & SQL
  queries, then a submitted fix of one or more statements & 5 & all test cases pass; reward is the fraction
  passed; feedback reports it and any error \\
\bottomrule
\end{tabular}
\end{table}

SWE-bench reference patches are never sent to rollout workers; BIRD reference
queries are available only to the grader; and ScienceWorld reference action
sequences are not generated during rollouts. The SWE-bench turn limit is reduced
from the configuration's default of 250 to 40. Attempts also have time limits of
1{,}800\,s in ScienceWorld and 7{,}200\,s in SWE-bench; three consecutive malformed
actions end a SWE-bench attempt. One ScienceWorld task,
\texttt{use-thermometer\_7}, has an annotated subgoal that cannot match the
environment's output, making full success unattainable on that task.

\subsection{Adaptation settings and implementation}
\label{app:configs}
\label{app:implementation}
\label{sec:implementation}
\label{app:protocol-constraints}

\begin{table}[!htbp]
\caption{\textbf{Online evaluation and context settings.}}
\label{tab:configs}
\centering
\small
\begin{tabular}{@{}>{\raggedright\arraybackslash}p{4.2cm}>{\raggedright\arraybackslash}p{8.8cm}@{}}
\toprule
Setting & Value \\
\midrule
Model & Qwen3.5-9B, non-thinking generation \\
Attempts per round $k$ & 16, sampled at temperature 1.0 \\
Rounds of attempts per task $m$ & 2 (frozen baseline, ICL-only and GRPO families); 3 (SDPO
  family) \\
Token allowance & task prompt 4{,}096; retained context 75{,}776; each attempt, including its
  observations, 24{,}576; teacher prompt 106{,}496; model sequence limit 131{,}072 \\
Turn limits & AppWorld (File System) 40; ScienceWorld (Measurement) 30; BIRD-Critic-SQLite (Financial) 5; SWE-bench Verified (Xarray) 40 \\
ICL context & replayed conversation with the agent's actions, observations, and the
  outcome's feedback; no limit on the number of episodes beyond the context allowance \\
\bottomrule
\end{tabular}
\end{table}

\begin{table}[!htbp]
\caption{\textbf{Weight-update settings.} Settings apply both with and without ICL.}
\label{tab:weight-update-settings}
\centering
\small
\begin{tabularx}{\linewidth}{@{}>{\raggedright\arraybackslash}p{5.0cm}>{\centering\arraybackslash}X>{\centering\arraybackslash}X@{}}
\toprule
Setting & GRPO & SDPO \\
\midrule
Adapter & LoRA & LoRA \\
Rank & 32 & 32 \\
LoRA $\alpha$ & 32 & 32 \\
Target layers & Linear except output head & Linear except output head \\
Optimizer & AdamW & AdamW \\
Learning rate & $10^{-5}$ & $10^{-5}$ \\
Learning-rate schedule & Constant & Constant \\
Weight decay & 0.01 & 0.01 \\
Gradient clipping & 1.0 & 1.0 \\
KL penalty toward base model & None & None \\
\midrule
Rounds per task & 2 & 3 \\
Epochs per round & 4 & 1 \\
Optimizer steps per round & 4 & 1 \\
Optimizer steps per task & 8 & 3 \\
Objective & Clipped policy gradient & Reverse KL \\
Loss aggregation & Token mean & Token mean, then attempt mean \\
Advantage centering & Round mean & --- \\
Advantage standard-deviation normalization & None & --- \\
Probability-ratio clip range & $[0.8,1.28]$ & --- \\
Dual clip (negative advantages) & 3 & --- \\
Importance-weight cap (each) & --- & 2 \\
Inference-to-trainer correction & None & Applied \\
\midrule
Self-teacher & --- & Exponential moving average \\
Teacher update rate & --- & 0.01 \\
KL support & --- & Student top 20 + remainder \\
Hindsight & --- & Sibling or own feedback \\
Hindsight donors per attempt & --- & At most 1 \\
\bottomrule
\end{tabularx}
\end{table}

Per task, GRPO receives 32 attempts and takes eight optimizer steps, whereas
SDPO receives 48 attempts and takes three steps. Our SDPO implementation supports
only one optimizer step per rollout batch, so we allocate an additional round to
provide more fresh on-policy experience and adaptation opportunities. The stateless
baseline and ICL-only mechanisms use the same two-round schedule as GRPO. SDPO's
extra round also changes context admission: \policy{All} can contribute up to three
episodes per task, compared with two for ICL-only and ICL$+$GRPO. Although SDPO
receives more interaction experience, GRPO takes more optimizer steps. Thus, these
comparisons concern complete adaptation procedures, not objectives at equal compute
or identical experience budgets.

\paragraph{Execution and reproducibility.}

Our framework extends verl v0.9.0 with a synchronous trainer and a common agent loop.
The stream is fixed before training, dataloader shuffling is disabled, and every
round is validated before any update or context admission. All expected attempts
must have matching identities, finite rewards, and valid endings; otherwise the
run stops. Failed tasks are not replaced. Updated weights are synchronized before
the next round, and all attempts within a round share context and weights.

Final grading outcomes become available only after an attempt ends; observations
from actions remain available during the attempt. An attempt
terminated by a turn, token, or time limit cannot succeed and receives zero
reward, except in ScienceWorld, where it retains partial subgoal reward.
Tolerated infrastructure faults are recorded as ungraded zero-reward outcomes;
a run is refused when at least four have occurred and they exceed 5\% of attempts.
A grading failure that leaves a submission without a verdict stops the run.
Oversized task prompts are rejected rather than silently truncated.

The self-teacher is a separate inference-only model in the actor worker.
Long-context scoring projects only the generated positions needed by the loss,
in chunks, avoiding a full sequence-by-vocabulary tensor. The chunked path was
checked against full-logit computation, and the SDPO objective and gradients
were checked against the reference implementation on CPU.

A run seed controls both stream order and generation seeds, although execution
is not bitwise reproducible. The recorded order is checked against the loaded
stream. Run configurations and input manifests record the model, benchmark,
source, and shared budget settings; the auditor checks these and compares logged
metrics with recorded outcomes. Best mechanisms in the main four-world summaries
are selected retrospectively by their five-seed mean success rate. During a run,
adaptation has access only to experience already generated, and each arriving
task is scored before learning from its outcomes.
\FloatBarrier

\section{Success Rates and Paired Comparisons}
\label{app:five-seed-results}
\label{app:metrics-formalism}

The 300 runs cover four worlds, fourteen adaptation mechanisms plus a stateless
baseline, and five seeds (0--4).
Table~\ref{tab:main} reports success rate using first-round outcomes, averaged
over tasks within a run before summarizing across seeds.

\subsection{Success rates by seed}
\label{app:success-by-seed}

Table~\ref{tab:main-seeds} expands Table~\ref{tab:main} into its five stream seeds.
Rows are grouped by seed to facilitate comparisons across mechanisms.
For each mechanism and seed, the mean weights the four worlds equally.

\begingroup
\small
\setlength{\tabcolsep}{2.5pt}
\setlength{\LTcapwidth}{\linewidth}
\newlength{\seedweightwidth}
\newlength{\seedpolicywidth}
\newlength{\seednumberwidth}
\settowidth{\seedweightwidth}{Weight update}
\settowidth{\seedpolicywidth}{\policy{Prefix\,\textperiodcentered\,Successful}}
\setlength{\seednumberwidth}{\dimexpr(\linewidth-\seedweightwidth-\seedpolicywidth-12\tabcolsep)/5\relax}
\begin{longtable}{@{}p{\seedweightwidth}p{\seedpolicywidth}*{5}{>{\centering\arraybackslash}p{\seednumberwidth}}@{}}
\caption{\textbf{Success rates by seed for all adaptation mechanisms and the baseline.}
Success rate (\%), separately for seeds 0--4. Mean weights the four worlds equally within each seed, before rounding.
Rows are grouped by seed; within each seed, mechanisms follow the grouping
and order of Table~\ref{tab:main}. Bold marks the highest value in each column
within each seed, including ties.}
\label{tab:main-seeds}\\
\toprule
Weight update & ICL policy & \shortstack{SWE-bench\\Verified\\{\scriptsize (Xarray)}} & \shortstack{AppWorld\\{\scriptsize (File System)}} & \shortstack{ScienceWorld\\{\scriptsize (Measurement)}} & \shortstack{BIRD-Critic-\\SQLite\\{\scriptsize (Financial)}} & Mean \\
\midrule
\endfirsthead
\caption[]{\textbf{Success rates by seed} (continued).}\\
\toprule
Weight update & ICL policy & \shortstack{SWE-bench\\Verified\\{\scriptsize (Xarray)}} & \shortstack{AppWorld\\{\scriptsize (File System)}} & \shortstack{ScienceWorld\\{\scriptsize (Measurement)}} & \shortstack{BIRD-Critic-\\SQLite\\{\scriptsize (Financial)}} & Mean \\
\midrule
\endhead
\midrule
\multicolumn{7}{r@{}}{\footnotesize Continued on next page}\\
\endfoot
\bottomrule
\endlastfoot
\multicolumn{7}{@{}l@{}}{%
\begin{tabular}[t]{@{}p{\seedweightwidth}p{\seedpolicywidth}*{5}{>{\centering\arraybackslash}p{\seednumberwidth}}@{}}
\multicolumn{7}{@{}l}{\textbf{Seed 0}}\\
\midrule
\rowcolor{mechanismFrozen}Frozen & \textsf{No ICL} & 11.6 & 56.5 & 17.6 & 29.7 & 28.9 \\
\midrule
\rowcolor{mechanismSDPO}SDPO & \textsf{No ICL} & 16.8 & 41.1 & 18.2 & 29.3 & 26.3 \\
\rowcolor{mechanismGRPO}GRPO & \textsf{No ICL} & 18.2 & 64.3 & 33.8 & 33.5 & 37.4 \\
\midrule
\rowcolor{mechanismICL}Frozen & \policy{FIFO\,\textperiodcentered\,All} & 35.8 & 74.4 & 48.6 & 27.1 & 46.5 \\
\rowcolor{mechanismICL} & \policy{FIFO\,\textperiodcentered\,Successful} & 32.1 & \textbf{87.8} & 54.8 & 32.8 & \textbf{51.9} \\
\rowcolor{mechanismICL} & \policy{Prefix\,\textperiodcentered\,All} & 35.5 & 76.8 & 47.7 & 27.4 & 46.9 \\
\rowcolor{mechanismICL} & \policy{Prefix\,\textperiodcentered\,Successful} & 33.0 & 82.4 & 54.0 & 37.2 & 51.6 \\
\midrule
\rowcolor{mechanismICLSDPO}SDPO & \policy{FIFO\,\textperiodcentered\,All} & 34.9 & 54.8 & 55.7 & 28.3 & 43.4 \\
\rowcolor{mechanismICLSDPO} & \policy{FIFO\,\textperiodcentered\,Successful} & 33.5 & 78.0 & 45.2 & 33.2 & 47.5 \\
\rowcolor{mechanismICLSDPO} & \policy{Prefix\,\textperiodcentered\,All} & 31.2 & 57.1 & \textbf{68.2} & 27.7 & 46.1 \\
\rowcolor{mechanismICLSDPO} & \policy{Prefix\,\textperiodcentered\,Successful} & 33.5 & 82.7 & 56.0 & 31.2 & 50.9 \\
\cmidrule(lr){1-7}
\rowcolor{mechanismICLGRPO}GRPO & \policy{FIFO\,\textperiodcentered\,All} & 37.2 & 69.0 & 47.4 & 34.5 & 47.1 \\
\rowcolor{mechanismICLGRPO} & \policy{FIFO\,\textperiodcentered\,Successful} & 35.8 & 74.4 & 57.7 & 33.2 & 50.3 \\
\rowcolor{mechanismICLGRPO} & \policy{Prefix\,\textperiodcentered\,All} & \textbf{39.8} & 77.1 & 48.9 & \textbf{39.5} & 51.3 \\
\rowcolor{mechanismICLGRPO} & \policy{Prefix\,\textperiodcentered\,Successful} & 34.9 & 79.8 & 58.0 & 31.5 & 51.0 \\
\end{tabular}}\\[1ex]
\multicolumn{7}{@{}l@{}}{%
\begin{tabular}[t]{@{}p{\seedweightwidth}p{\seedpolicywidth}*{5}{>{\centering\arraybackslash}p{\seednumberwidth}}@{}}
\multicolumn{7}{@{}l}{\textbf{Seed 1}}\\
\midrule
\rowcolor{mechanismFrozen}Frozen & \textsf{No ICL} & 14.5 & 57.4 & 16.2 & 30.1 & 29.6 \\
\midrule
\rowcolor{mechanismSDPO}SDPO & \textsf{No ICL} & 18.8 & 46.1 & 17.3 & 26.4 & 27.2 \\
\rowcolor{mechanismGRPO}GRPO & \textsf{No ICL} & 21.0 & 62.5 & 30.1 & 36.2 & 37.5 \\
\midrule
\rowcolor{mechanismICL}Frozen & \policy{FIFO\,\textperiodcentered\,All} & 42.6 & 56.2 & 53.1 & 30.8 & 45.7 \\
\rowcolor{mechanismICL} & \policy{FIFO\,\textperiodcentered\,Successful} & 34.9 & 79.2 & 43.2 & 35.5 & 48.2 \\
\rowcolor{mechanismICL} & \policy{Prefix\,\textperiodcentered\,All} & 37.5 & 66.4 & 48.6 & 26.1 & 44.6 \\
\rowcolor{mechanismICL} & \policy{Prefix\,\textperiodcentered\,Successful} & 32.1 & \textbf{81.2} & 56.2 & 31.5 & 50.3 \\
\midrule
\rowcolor{mechanismICLSDPO}SDPO & \policy{FIFO\,\textperiodcentered\,All} & 40.6 & 49.1 & 58.0 & 29.3 & 44.2 \\
\rowcolor{mechanismICLSDPO} & \policy{FIFO\,\textperiodcentered\,Successful} & 39.2 & \textbf{81.2} & 64.5 & 29.0 & \textbf{53.5} \\
\rowcolor{mechanismICLSDPO} & \policy{Prefix\,\textperiodcentered\,All} & 41.8 & 47.3 & \textbf{64.8} & 34.1 & 47.0 \\
\rowcolor{mechanismICLSDPO} & \policy{Prefix\,\textperiodcentered\,Successful} & 38.9 & 75.9 & 54.3 & 33.1 & 50.5 \\
\cmidrule(lr){1-7}
\rowcolor{mechanismICLGRPO}GRPO & \policy{FIFO\,\textperiodcentered\,All} & 37.8 & 62.8 & 58.2 & 39.1 & 49.5 \\
\rowcolor{mechanismICLGRPO} & \policy{FIFO\,\textperiodcentered\,Successful} & 40.6 & 80.1 & 51.4 & 32.8 & 51.2 \\
\rowcolor{mechanismICLGRPO} & \policy{Prefix\,\textperiodcentered\,All} & \textbf{43.2} & 71.4 & 54.0 & 35.1 & 50.9 \\
\rowcolor{mechanismICLGRPO} & \policy{Prefix\,\textperiodcentered\,Successful} & 42.0 & 69.0 & 58.8 & \textbf{41.3} & 52.8 \\
\end{tabular}}\\[1ex]
\multicolumn{7}{@{}l@{}}{%
\begin{tabular}[t]{@{}p{\seedweightwidth}p{\seedpolicywidth}*{5}{>{\centering\arraybackslash}p{\seednumberwidth}}@{}}
\multicolumn{7}{@{}l}{\textbf{Seed 2}}\\
\midrule
\rowcolor{mechanismFrozen}Frozen & \textsf{No ICL} & 16.5 & 57.4 & 15.3 & 28.6 & 29.5 \\
\midrule
\rowcolor{mechanismSDPO}SDPO & \textsf{No ICL} & 15.9 & 46.7 & 20.5 & 29.0 & 28.0 \\
\rowcolor{mechanismGRPO}GRPO & \textsf{No ICL} & 21.0 & 56.5 & 28.7 & 31.7 & 34.5 \\
\midrule
\rowcolor{mechanismICL}Frozen & \policy{FIFO\,\textperiodcentered\,All} & 36.6 & 62.5 & 43.8 & 27.1 & 42.5 \\
\rowcolor{mechanismICL} & \policy{FIFO\,\textperiodcentered\,Successful} & 34.4 & 70.8 & 24.4 & 30.8 & 40.1 \\
\rowcolor{mechanismICL} & \policy{Prefix\,\textperiodcentered\,All} & 35.8 & 46.4 & 34.9 & 26.8 & 36.0 \\
\rowcolor{mechanismICL} & \policy{Prefix\,\textperiodcentered\,Successful} & 32.4 & 73.2 & 46.0 & 33.9 & 46.4 \\
\midrule
\rowcolor{mechanismICLSDPO}SDPO & \policy{FIFO\,\textperiodcentered\,All} & 39.8 & 57.4 & \textbf{54.0} & 30.8 & 45.5 \\
\rowcolor{mechanismICLSDPO} & \policy{FIFO\,\textperiodcentered\,Successful} & 39.2 & \textbf{81.0} & 31.0 & 28.8 & 45.0 \\
\rowcolor{mechanismICLSDPO} & \policy{Prefix\,\textperiodcentered\,All} & 38.4 & 56.5 & 40.9 & 29.0 & 41.2 \\
\rowcolor{mechanismICLSDPO} & \policy{Prefix\,\textperiodcentered\,Successful} & 32.1 & 74.7 & 45.7 & 29.3 & 45.5 \\
\cmidrule(lr){1-7}
\rowcolor{mechanismICLGRPO}GRPO & \policy{FIFO\,\textperiodcentered\,All} & \textbf{45.7} & 70.2 & 45.5 & 30.4 & \textbf{48.0} \\
\rowcolor{mechanismICLGRPO} & \policy{FIFO\,\textperiodcentered\,Successful} & 34.1 & 72.9 & 50.3 & 34.1 & 47.8 \\
\rowcolor{mechanismICLGRPO} & \policy{Prefix\,\textperiodcentered\,All} & 40.9 & 59.5 & 53.1 & 29.8 & 45.8 \\
\rowcolor{mechanismICLGRPO} & \policy{Prefix\,\textperiodcentered\,Successful} & 36.1 & 61.3 & 53.7 & \textbf{35.9} & 46.8 \\
\end{tabular}}\\[1ex]
\multicolumn{7}{@{}l@{}}{%
\begin{tabular}[t]{@{}p{\seedweightwidth}p{\seedpolicywidth}*{5}{>{\centering\arraybackslash}p{\seednumberwidth}}@{}}
\multicolumn{7}{@{}l}{\textbf{Seed 3}}\\
\midrule
\rowcolor{mechanismFrozen}Frozen & \textsf{No ICL} & 10.8 & 56.2 & 15.6 & 26.6 & 27.3 \\
\midrule
\rowcolor{mechanismSDPO}SDPO & \textsf{No ICL} & 18.2 & 42.9 & 20.2 & 27.3 & 27.1 \\
\rowcolor{mechanismGRPO}GRPO & \textsf{No ICL} & 26.1 & 60.4 & 37.2 & 29.7 & 38.4 \\
\midrule
\rowcolor{mechanismICL}Frozen & \policy{FIFO\,\textperiodcentered\,All} & 37.2 & 70.8 & 57.7 & 29.3 & 48.7 \\
\rowcolor{mechanismICL} & \policy{FIFO\,\textperiodcentered\,Successful} & 36.6 & 72.3 & 50.9 & 32.8 & 48.2 \\
\rowcolor{mechanismICL} & \policy{Prefix\,\textperiodcentered\,All} & 40.3 & 58.6 & 55.7 & 29.0 & 45.9 \\
\rowcolor{mechanismICL} & \policy{Prefix\,\textperiodcentered\,Successful} & 34.7 & 67.6 & 54.8 & 33.1 & 47.5 \\
\midrule
\rowcolor{mechanismICLSDPO}SDPO & \policy{FIFO\,\textperiodcentered\,All} & 36.4 & 40.8 & 62.8 & 34.8 & 43.7 \\
\rowcolor{mechanismICLSDPO} & \policy{FIFO\,\textperiodcentered\,Successful} & 37.8 & 62.5 & 60.5 & 33.8 & 48.7 \\
\rowcolor{mechanismICLSDPO} & \policy{Prefix\,\textperiodcentered\,All} & 37.8 & 65.2 & 48.3 & 27.8 & 44.8 \\
\rowcolor{mechanismICLSDPO} & \policy{Prefix\,\textperiodcentered\,Successful} & 38.4 & 76.5 & 59.7 & 28.8 & 50.8 \\
\cmidrule(lr){1-7}
\rowcolor{mechanismICLGRPO}GRPO & \policy{FIFO\,\textperiodcentered\,All} & 37.8 & 64.0 & 46.0 & \textbf{38.9} & 46.7 \\
\rowcolor{mechanismICLGRPO} & \policy{FIFO\,\textperiodcentered\,Successful} & 31.5 & 77.1 & 63.4 & 31.5 & 50.9 \\
\rowcolor{mechanismICLGRPO} & \policy{Prefix\,\textperiodcentered\,All} & \textbf{43.2} & 62.8 & \textbf{63.6} & 32.2 & 50.5 \\
\rowcolor{mechanismICLGRPO} & \policy{Prefix\,\textperiodcentered\,Successful} & 39.8 & \textbf{79.2} & 61.1 & 36.6 & \textbf{54.2} \\
\end{tabular}}\\[1ex]
\multicolumn{7}{@{}l@{}}{%
\begin{tabular}[t]{@{}p{\seedweightwidth}p{\seedpolicywidth}*{5}{>{\centering\arraybackslash}p{\seednumberwidth}}@{}}
\multicolumn{7}{@{}l}{\textbf{Seed 4}}\\
\midrule
\rowcolor{mechanismFrozen}Frozen & \textsf{No ICL} & 13.4 & 60.7 & 13.4 & 27.4 & 28.7 \\
\midrule
\rowcolor{mechanismSDPO}SDPO & \textsf{No ICL} & 14.2 & 47.3 & 19.9 & 26.8 & 27.1 \\
\rowcolor{mechanismGRPO}GRPO & \textsf{No ICL} & 22.4 & 62.8 & 29.0 & 34.5 & 37.2 \\
\midrule
\rowcolor{mechanismICL}Frozen & \policy{FIFO\,\textperiodcentered\,All} & \textbf{44.6} & 70.5 & 48.3 & 28.3 & 47.9 \\
\rowcolor{mechanismICL} & \policy{FIFO\,\textperiodcentered\,Successful} & 37.5 & 79.5 & 57.1 & 30.8 & 51.2 \\
\rowcolor{mechanismICL} & \policy{Prefix\,\textperiodcentered\,All} & 29.0 & 64.6 & 52.8 & 30.0 & 44.1 \\
\rowcolor{mechanismICL} & \policy{Prefix\,\textperiodcentered\,Successful} & 34.9 & \textbf{86.3} & 55.1 & 27.8 & 51.1 \\
\midrule
\rowcolor{mechanismICLSDPO}SDPO & \policy{FIFO\,\textperiodcentered\,All} & 40.1 & 66.4 & 52.6 & 32.4 & 47.8 \\
\rowcolor{mechanismICLSDPO} & \policy{FIFO\,\textperiodcentered\,Successful} & 31.2 & 73.5 & \textbf{62.8} & 34.4 & 50.5 \\
\rowcolor{mechanismICLSDPO} & \policy{Prefix\,\textperiodcentered\,All} & 19.3 & 68.8 & 55.4 & 30.3 & 43.4 \\
\rowcolor{mechanismICLSDPO} & \policy{Prefix\,\textperiodcentered\,Successful} & 33.0 & 74.4 & 35.5 & 30.7 & 43.4 \\
\cmidrule(lr){1-7}
\rowcolor{mechanismICLGRPO}GRPO & \policy{FIFO\,\textperiodcentered\,All} & 35.8 & 64.6 & 53.1 & 29.3 & 45.7 \\
\rowcolor{mechanismICLGRPO} & \policy{FIFO\,\textperiodcentered\,Successful} & 40.9 & 77.4 & 53.4 & 33.1 & 51.2 \\
\rowcolor{mechanismICLGRPO} & \policy{Prefix\,\textperiodcentered\,All} & 32.7 & 76.2 & 59.9 & 33.4 & 50.5 \\
\rowcolor{mechanismICLGRPO} & \policy{Prefix\,\textperiodcentered\,Successful} & 34.9 & 83.3 & 59.1 & \textbf{36.1} & \textbf{53.4} \\
\end{tabular}}\\[1ex]
\end{longtable}
\endgroup

\FloatBarrier
\subsection{Complete success-rate gains and wins}
\label{app:complete-stream-comparisons}
\label{app:sdpo-stream-comparisons}

\begin{figure}[!htbp]
\centering
\includegraphics[width=\linewidth]{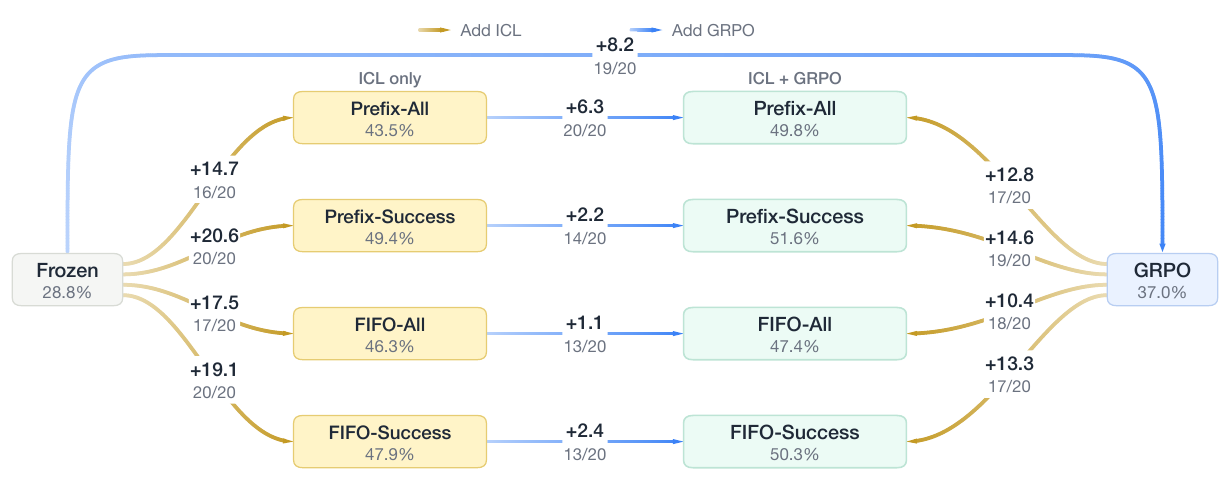}
\caption{\textbf{Mean gains and wins from adding ICL and GRPO.}
Boxes show mean success rates. Each arrow compares its destination with its
source: yellow adds ICL, blue adds GRPO. Arrow labels show the mean success-rate gain in percentage points (top) and wins
out of 20 world--seed stream pairs (bottom), weighting all streams equally.}
\label{fig:stream-win-rates-grpo-full}
\end{figure}

\begin{figure}[!htbp]
\centering
\includegraphics[width=\linewidth]{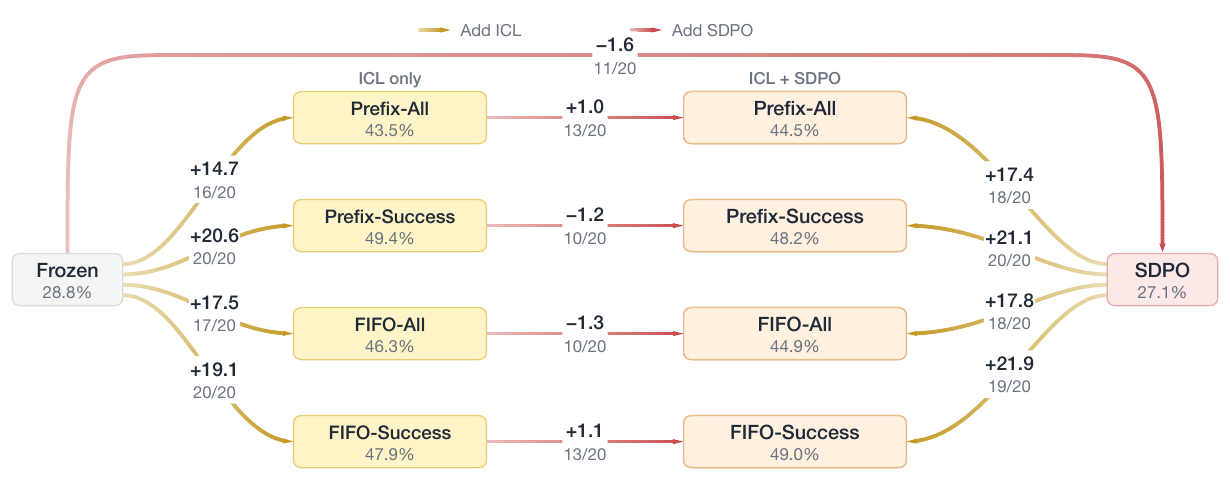}
\caption{\textbf{Mean gains and wins from adding ICL and SDPO.}
Boxes show mean success rates. Each arrow compares its destination with its
source: yellow adds ICL, red adds SDPO. Arrow labels show the mean success-rate gain in percentage points (top) and wins
out of 20 world--seed stream pairs (bottom), weighting all streams equally.}
\label{fig:stream-win-rates-sdpo}
\end{figure}

Figures~\ref{fig:stream-win-rates-grpo-full}
and~\ref{fig:stream-win-rates-sdpo} expand the GRPO and SDPO comparisons in
Figure~\ref{fig:stream-win-rates}, using the same 20 world--seed pairs.
They additionally show gains from adding ICL to Frozen and to each
weight-only mechanism.

\FloatBarrier

\section{Interaction Length and Action Composition}
\label{app:additional-behavior}
\label{app:execution-detail}
\label{app:turns}

These analyses complement Sections~\ref{sec:single-channel}
and~\ref{sec:hybrid-mechanisms}. We separate interaction length from the
number of operations written within an action.

\subsection{Interaction length by outcome}
\label{app:turns-by-outcome}

\begin{figure}[!htbp]
\centering
\includegraphics[width=\linewidth]{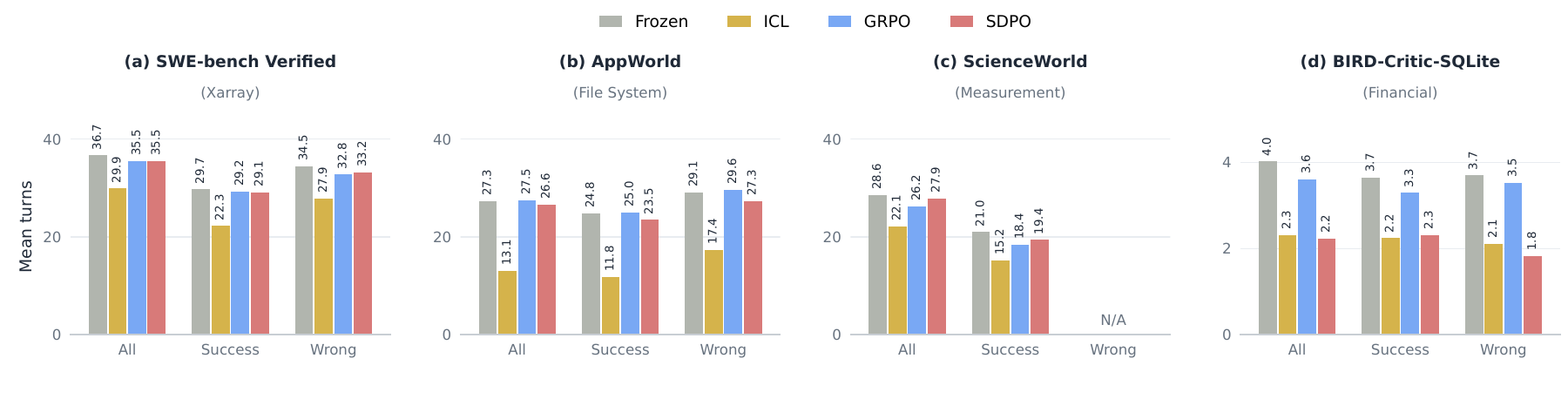}
\caption{\textbf{Successful attempts are also shorter with ICL.}
Mean turns over all attempts (All), successful attempts (Success), and submitted
but unsuccessful attempts (Wrong), averaged equally over five seeds.
ScienceWorld has no separate Wrong category (N/A). These are full-stream means,
not temporal trends; policies and conditioning are defined above.}
\label{fig:single-channel-conditional-turns}
\end{figure}

Figure~\ref{fig:single-channel-conditional-turns} separates mean attempt
length by outcome for the single-channel mechanisms. ICL uses fewer turns than
the frozen baseline even among successful attempts in each benchmark, so its
lower overall mean is not solely due to a smaller share of unfinished attempts.

\paragraph{Measurement.}
We average turns over eligible attempts within each stream seed, then weight
the five seeds equally. ICL uses \policy{FIFO}--\policy{All} on SWE-bench and
\policy{Prefix}--\policy{Successful} elsewhere, as in
Figure~\ref{fig:single-channel-outcomes-turns}; GRPO and SDPO use no ICL.
ScienceWorld has no separate wrong-submission outcome.

\clearpage
\subsection{Multiple operations per action}
\label{app:operations-per-action}

\begin{figure}[!htbp]
\centering
\includegraphics[width=\linewidth]{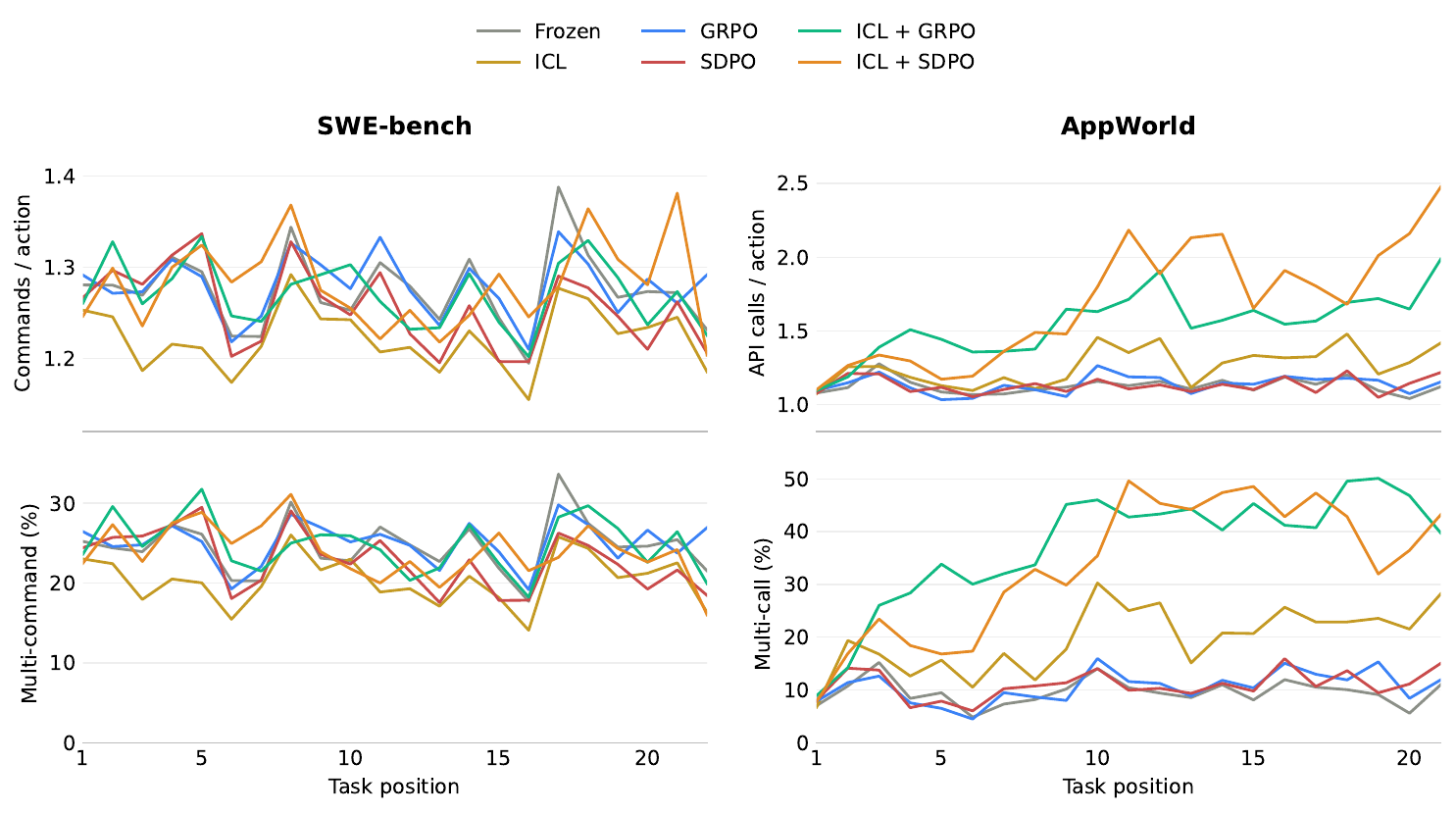}
\caption{\textbf{Multiple operations can share one action.}
Top: shell commands per action in SWE-bench and API call sites per action in
AppWorld. Bottom: percentage of eligible actions containing at least two such
operations. Curves show unsmoothed means over five seeds using first-round
attempts. Counts describe written operations; measurement rules are given above.}
\label{fig:action-packing-appendix}
\end{figure}

Figure~\ref{fig:action-packing-appendix} complements the turn counts by measuring
how many operations are written within one action in SWE-bench and AppWorld.
These measures describe written action structure, not executed operation counts,
token cost, or wall-clock cost.

\paragraph{Measurement.}
In SWE-bench, we count separate shell invocations, including pipeline components,
but exclude setup-only commands and final submissions; embedded Python counts
as one invocation. In AppWorld, we count explicit
\texttt{apis.<app>.<endpoint>(...)} call sites, including documentation and
supervisor calls. Calls inside loops count once as written, not once per
execution. Refused and unparseable actions are excluded. We average eligible
actions within each attempt, attempts within each seed and task position, then
the five seeds equally. ICL and its hybrids use \policy{FIFO}--\policy{All} on
SWE-bench and \policy{Prefix}--\policy{Successful} on AppWorld.

\FloatBarrier

\section{Context Retention and Weight Adaptation}
\label{app:retention-focus}
\label{app:memory-capacity}

This appendix examines how admission and retention policies shape the available
context and the gains from weight updates. We first report gains and context
occupancy, then compare retention policies, and finally distinguish reward
variation from the admission of new experience.

\subsection{Cumulative gains and tasks in context}
\label{app:context-gains}

\paragraph{Measurement.}
Cumulative gain is the running sum of success-rate differences, in percentage
points, through each task position. Differences pair the same task and seed,
using 16 attempts per mechanism; we sum within each seed, then average the five
seeds equally. Context occupancy counts distinct task IDs after budget fitting,
before each task begins; different task variants count separately. Counts also
average the five seeds equally. Colors distinguish
\policy{FIFO} and \policy{Prefix}; solid lines denote \policy{All} and dashed
lines \policy{Successful}.

\begin{figure}[!htbp]
\centering
\includegraphics[width=\linewidth]{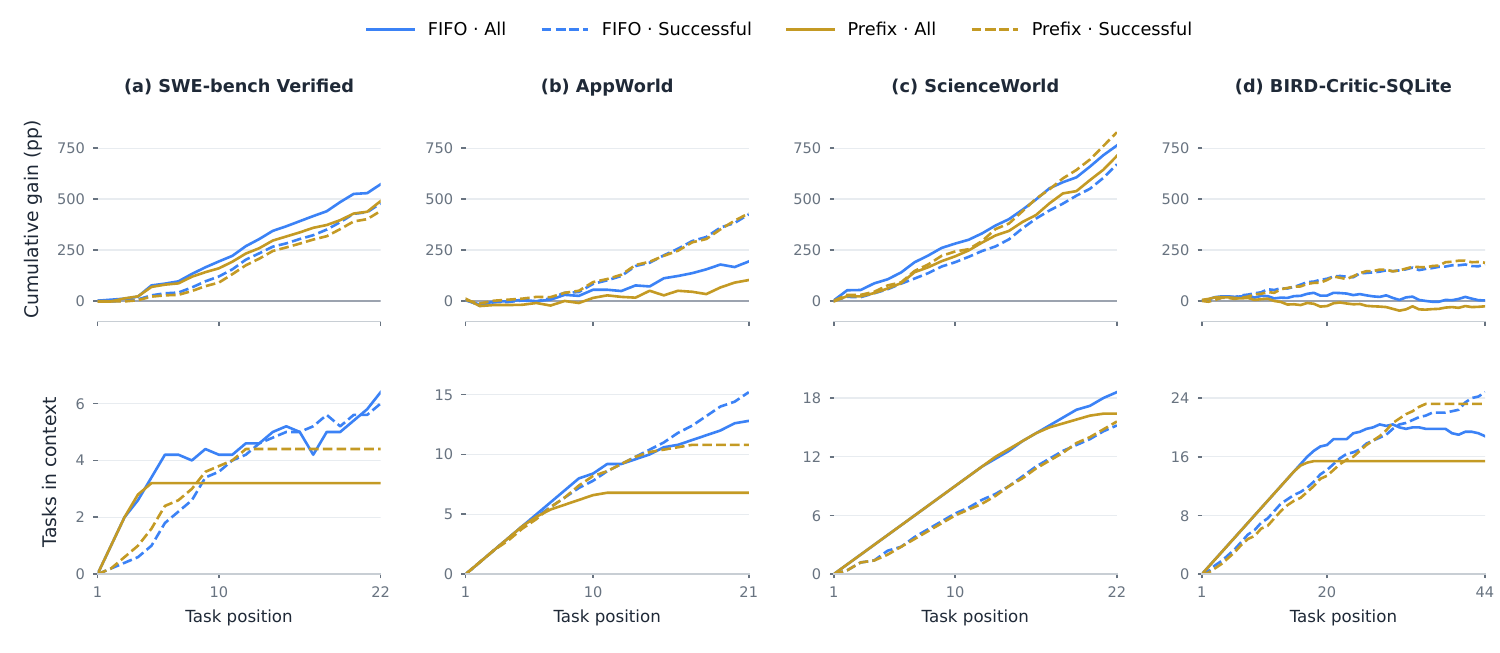}
\caption{\textbf{Cumulative ICL gains and retained context.}
Top: cumulative success-rate gain over Frozen, in percentage points.
Bottom: distinct tasks in context before each task begins. All weights are
frozen. Curves show means over five seeds; pairing and context counts are
defined above.}
\label{fig:icl-cumulative-gain-context}
\end{figure}

Figure~\ref{fig:icl-cumulative-gain-context} compares all four
frozen-weight ICL policies over the stream. It complements the representative
policy used for each benchmark in Section~\ref{sec:icl-effects}, showing both
cumulative gains over the no-ICL baseline and the tasks represented in
retained context.

\begin{figure}[!htbp]
\centering
\includegraphics[width=\linewidth]{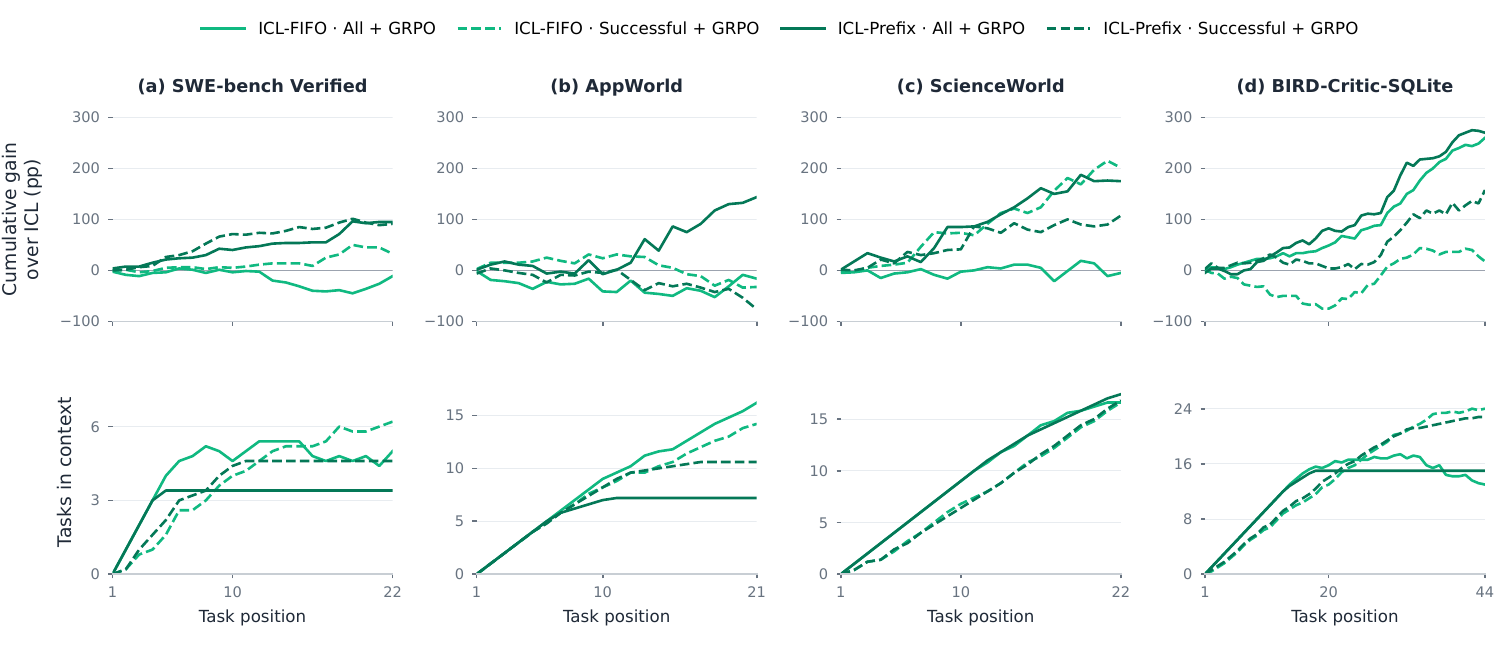}
\caption{\textbf{Adding GRPO to ICL: cumulative gains and retained context.}
Top: cumulative success-rate gain over the matching frozen-weight ICL policy,
in percentage points. Bottom: distinct tasks in the hybrid's context before
each task begins. Curves show means over five seeds, using the pairing and
context counts defined above.}
\label{fig:hybrid-grpo-context}
\label{fig:hybrid-context}
\end{figure}

\begin{figure}[!htbp]
\centering
\includegraphics[width=\linewidth]{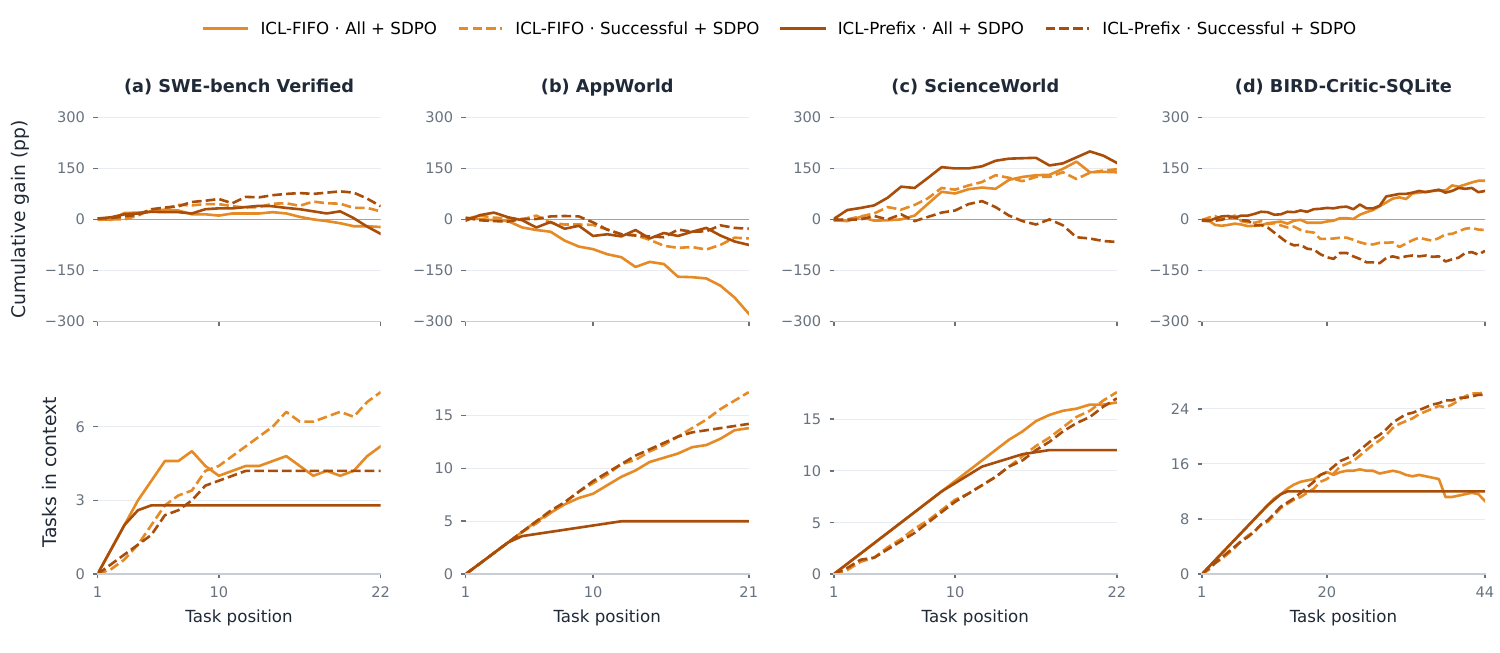}
\caption{\textbf{Adding SDPO to ICL: cumulative gains and retained context.}
Top: cumulative success-rate gain over the matching frozen-weight ICL policy,
in percentage points. Bottom: distinct tasks in the hybrid's context before
each task begins. Curves show means over five seeds, using the pairing and
context counts defined above.}
\label{fig:hybrid-sdpo-context}
\end{figure}

Figures~\ref{fig:hybrid-grpo-context} and~\ref{fig:hybrid-sdpo-context}
extend the hybrid analysis in Section~\ref{sec:hybrid-mechanisms} to all four
ICL policies, showing cumulative gains from GRPO and SDPO over each
frozen-weight ICL policy alongside the tasks represented in the hybrid's
context.

\FloatBarrier
\subsection{Prefix versus FIFO retention}
\label{app:retention-comparisons}

Table~\ref{tab:retention-world-selection} summarizes explicit history under the
75,776-token allowance, with weights fixed and \policy{All} admission. \policy{FIFO} evicts old
episodes; \policy{Prefix} stops admission at overflow.

\begin{table}[!htbp]
\centering\small
\caption{\textbf{Context capacity, weights fixed and \policy{All} admission.} Five-seed
means. Evictions: preparations evicting episodes. Freeze: first task position
where admission stops. Retained: distinct prior tasks before the final task's
first round; Available: total prior tasks.}
\label{tab:retention-world-selection}
\begin{tabular}{lrrrr}
\toprule
Benchmark & \policy{FIFO} evictions (\%) & \policy{Prefix} freeze & Retained (\policy{FIFO} / \policy{Prefix}) & Available \\
\midrule
AppWorld (File System) & 31.0 & 7.8 & 12.8 / 6.8 & 20 \\
ScienceWorld (Measurement) & 11.4 & 17.4 & 18.6 / 16.4 & 21 \\
BIRD-Critic-SQLite (Financial) & 46.4 & 16.4 & 18.8 / 15.4 & 43 \\
SWE-bench Verified (Xarray) & 47.7 & 4.2 & 6.4 / 3.2 & 21 \\
\bottomrule
\end{tabular}
\end{table}

Reconstructed task histories are checked against telemetry.
Under \policy{Successful} admission, which limits contributions
to one episode per task, ScienceWorld with fixed weights never evicts or freezes.

\paragraph{Performance after admission stops.}
On SWE-bench with \policy{All} admission, all \policy{Prefix} runs with weights fixed or GRPO freeze
by position 5. Over positions 6--22, GRPO raises the \policy{Prefix} success rate from 39.2\% to
43.5\% ($+4.3\pm2.4$ points); \policy{FIFO} gains $-0.4\pm8.0$ points. The window is chosen
from telemetry, not performance. SDPO does not reproduce the \policy{Prefix} success-rate gain.

\paragraph{Interpretation.}
Runs develop different policies and retained contexts. These comparisons do not establish
preservation of excluded episodes or more stable optimization. Post-freeze gains
can reflect earlier updates; isolating subsequent learning requires a control
that freezes weights at the same cutoff. Paired SDs measure seed variation,
not temporal stability.
\FloatBarrier

\subsection{Reward variation and context admission}
\label{app:reward-context-admission}

\begin{figure}[!htbp]
\centering
\includegraphics[width=\linewidth]{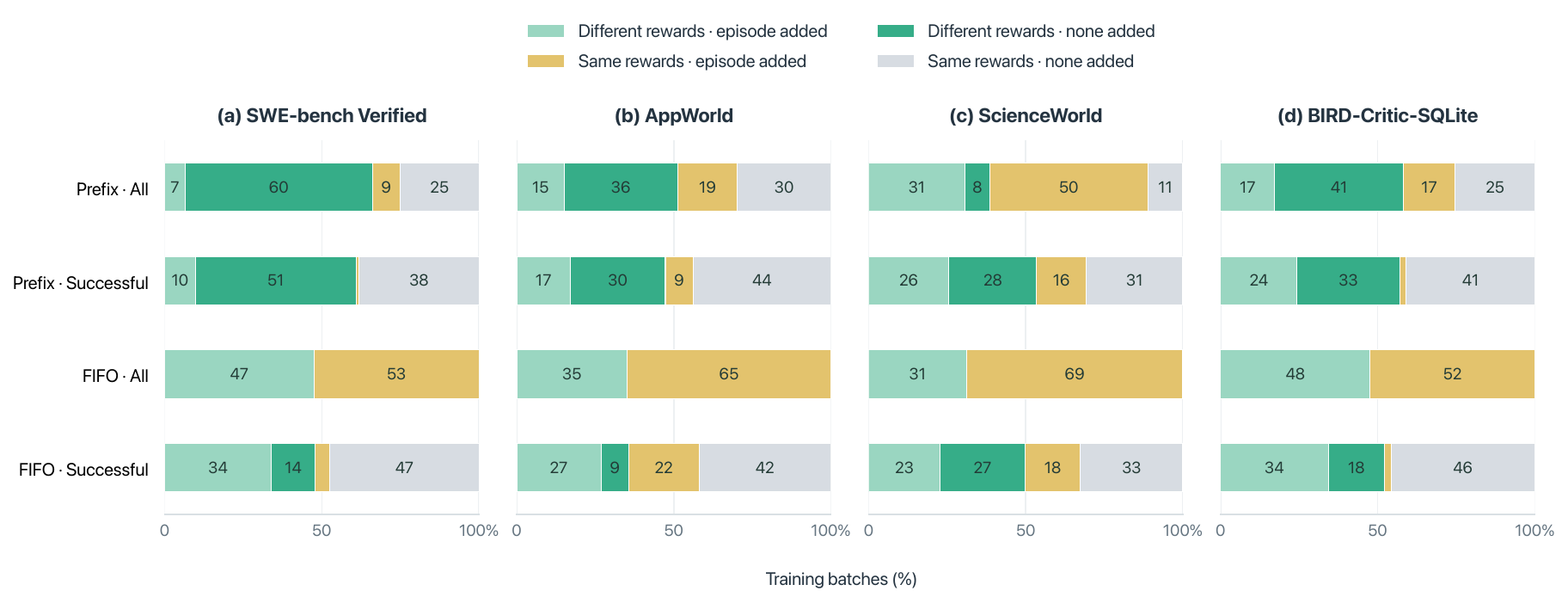}
\caption{\textbf{Reward variation and context admission with GRPO.}
Each bar partitions training batches by reward variation and admission of a
new context episode for one ICL policy. Percentages pool batches across five
seeds; category definitions are given above.}
\label{fig:grpo-reward-context-admission}
\end{figure}

\begin{figure}[!htbp]
\centering
\includegraphics[width=\linewidth]{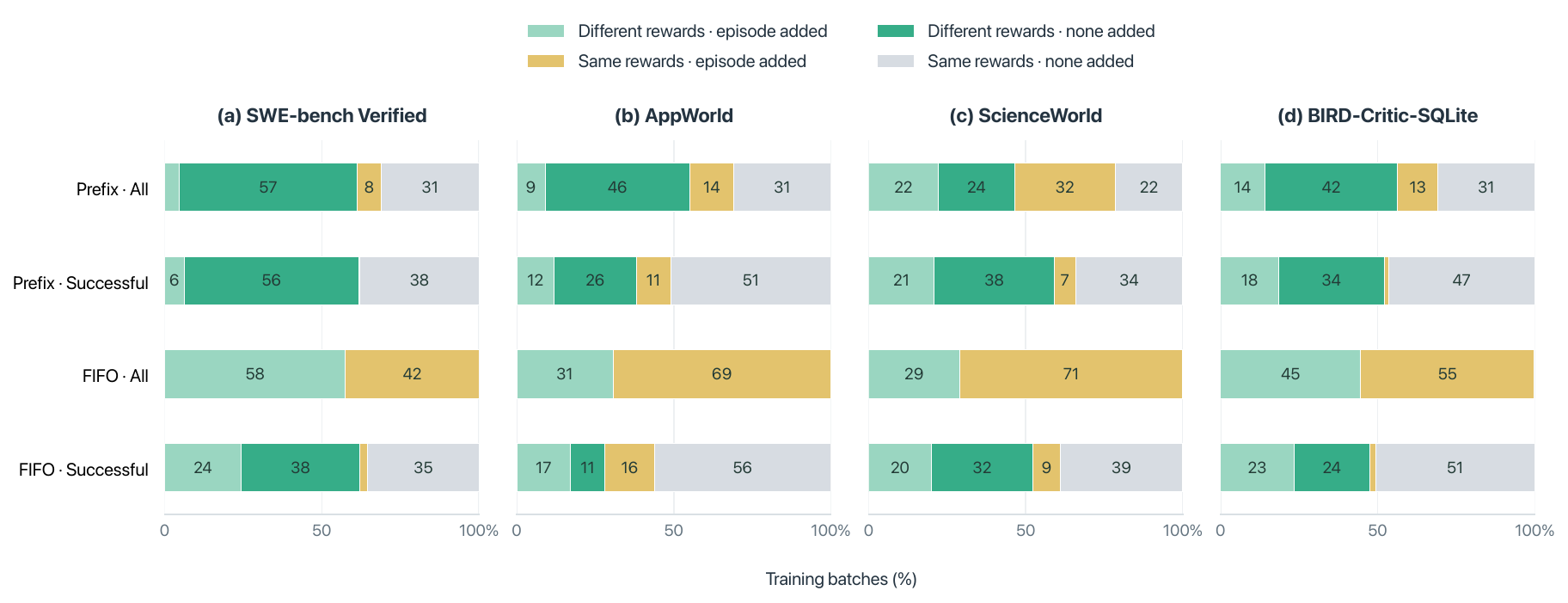}
\caption{\textbf{Reward variation and context admission with SDPO.}
Each bar partitions training batches by reward variation and admission of a
new context episode for one ICL policy. Percentages pool batches across five
seeds; category definitions are given above.}
\label{fig:sdpo-reward-context-admission}
\end{figure}

Figures~\ref{fig:grpo-reward-context-admission} and~\ref{fig:sdpo-reward-context-admission}
separate reward variation within a training batch from admission of an episode
into context. These are distinct properties: a batch can supply a GRPO reward
contrast without contributing a new demonstration. For SDPO, reward variation
is descriptive rather than a requirement for distillation supervision.

\paragraph{Measurement.}
Each training batch contains 16 attempts. Rewards vary when their maximum and
minimum differ by more than $10^{-7}$. An episode is counted as admitted when
it appears in the next prepared context after budget fitting, accounting for
\policy{FIFO} evictions. Percentages pool batches with recorded admission
outcomes across five seeds and are rounded for display. A reward contrast
without a new context episode measures an available GRPO training signal,
not its contribution to later success. SDPO supervision can remain informative
even when all attempts receive the same reward; reward variation does not
measure the presence or absence of its learning signal.

\FloatBarrier

\clearpage
\section{Additional Environments}
\label{app:additional-worlds}

We additionally evaluate 32 scikit-learn tasks from SWE-bench Verified and
64 Toxicology tasks from BIRD-CRITIC, using Qwen3.5-9B, seed~4, and
16 attempts per round. We compare the frozen baseline, ICL, GRPO, SDPO,
and the two hybrids. All ICL mechanisms use
\policy{Prefix}--\policy{Successful}. As in the main online evaluation,
we score the first round of each task before adapting to its outcomes.
These runs use one seed and do not estimate variability across task orders.

\begin{figure}[!htbp]
\centering
\includegraphics[width=\linewidth]{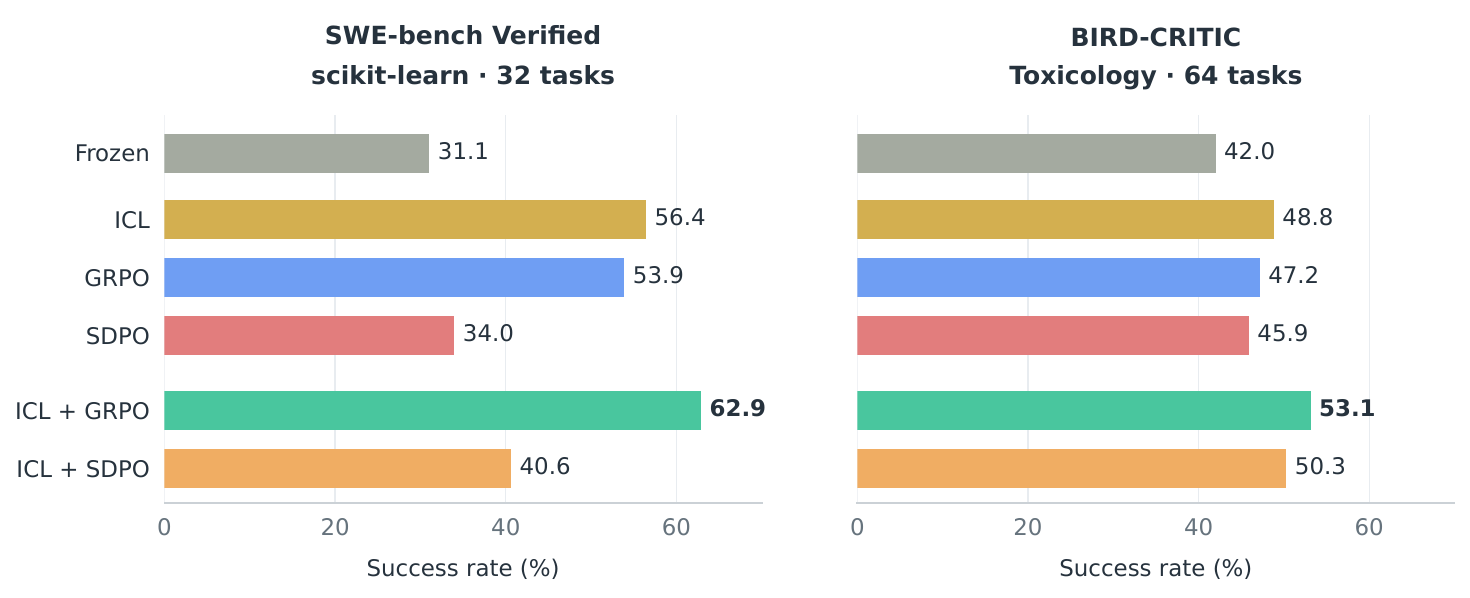}
\caption{\textbf{Success rates in two additional environments.}
Results use seed~4 and 16 attempts per task, with
\policy{Prefix}--\policy{Successful} for all ICL mechanisms.
Bold labels mark the highest success rate in each environment.}
\label{fig:additional-worlds-success}
\end{figure}

\begin{figure}[!htbp]
\centering
\includegraphics[width=\linewidth]{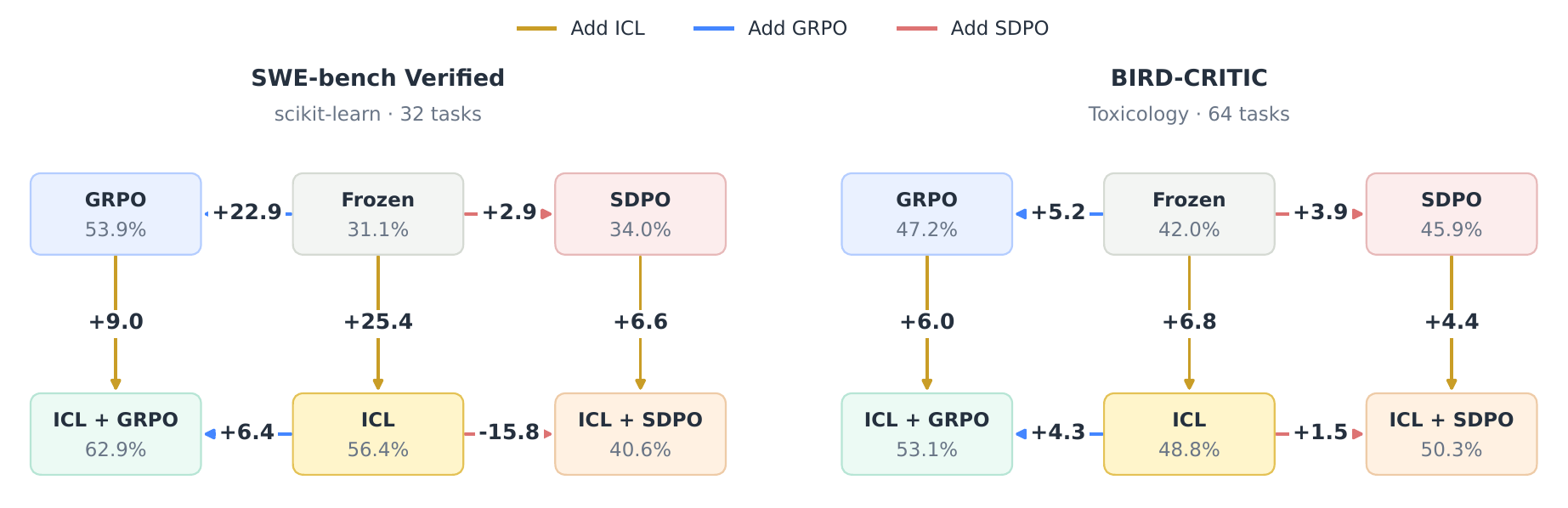}
\caption{\textbf{Gains from adding context and weight adaptation.}
Boxes show success rates; arrows show the change in percentage points
from source to destination for the experiments in
Figure~\ref{fig:additional-worlds-success}.}
\label{fig:additional-worlds-gains}
\end{figure}

ICL$+$GRPO achieves the highest success rate among the six tested mechanisms
in both environments (Figure~\ref{fig:additional-worlds-success}). Adding
GRPO improves ICL by 6.4 percentage points on scikit-learn and 4.3 on
Toxicology. SDPO's incremental effect remains environment-dependent:
adding it to ICL lowers success on scikit-learn but improves it on
Toxicology (Figure~\ref{fig:additional-worlds-gains}). These results extend
the comparison to two larger worlds and support the main finding that
GRPO can complement retained experience, while the benefit of combining
channels depends on the weight-update mechanism.

\FloatBarrier

\clearpage
\section{Transfer and Component Ablations}
\label{app:transfer}

We evaluate the final seed-4 agents obtained with
\policy{Prefix}--\policy{All}$+$GRPO and
\policy{Prefix}--\policy{Successful}$+$GRPO on each of the four worlds.
For each source world, we evaluate three versions of the same hybrid agent:
\emph{both}, retaining its adapted weights and memory;
\emph{adapter only}, removing its retained memory; and
\emph{memory only}, using its retained memory with the original model weights.
Memory only therefore differs from an independently adapted ICL-only agent:
it uses the experience collected by the hybrid.
Evaluation keeps the transferred state fixed and uses 16 attempts per task,
with the same target-task order across mechanisms. Success rates average
over all tasks in the target world.

The two policies were evaluated in separate rounds, each with a fresh frozen,
no-ICL baseline in every target world. We compare each result with its own
round's baseline; for example, the BIRD baselines are 30.8\% and 26.8\%.
These evaluations use one adapted agent per policy and source world,
so they do not measure variation across adaptation seeds.

\subsection{Component ablations on the same encountered tasks}
\label{app:transfer-components}

\begin{figure}[!htbp]
\centering
\includegraphics[width=\linewidth]{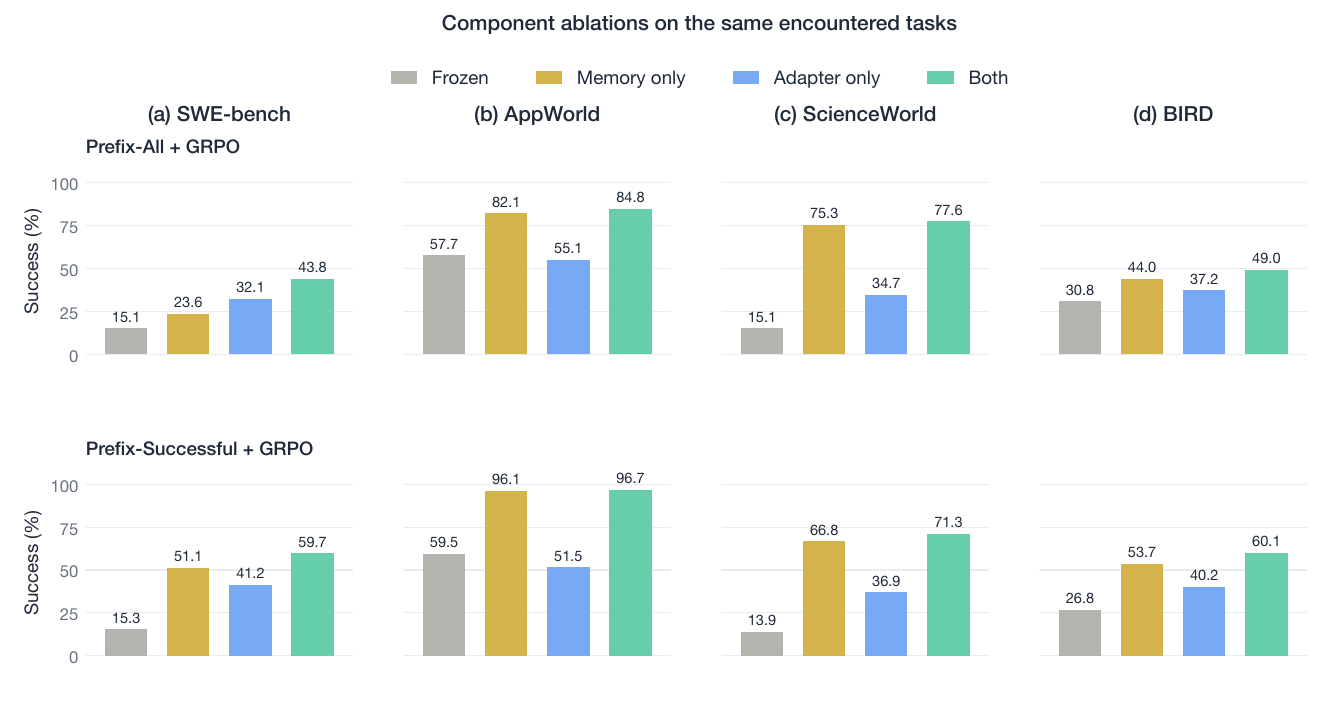}
\caption{\textbf{Component ablations on the same encountered tasks.}
Success rates of final seed-4 hybrid agents on their original worlds.
Both retains the adapted weights and memory; the other two variants remove
one component. Frozen is the fresh baseline from the corresponding round.
Each bar averages 16 attempts per task. These are previously encountered
tasks, not held-out evaluations.}
\label{fig:transfer-components}
\end{figure}

Figure~\ref{fig:transfer-components} evaluates each agent on its original
world's tasks. These tasks were encountered during adaptation, so the
results can reflect task-specific memorization and do not establish
held-out generalization. They instead show how removing either component
changes the final agent's performance on those tasks.

The full hybrid has a higher success rate than either component alone in
all eight policy--world comparisons. The adapter alone remains above the
baseline on SWE-bench, ScienceWorld, and BIRD under both policies, but falls
below it on AppWorld. The relative contributions differ across worlds:
for \policy{Prefix}--\policy{All} on SWE-bench, both components yield
43.8\% success, compared with 32.1\% for the adapter alone and 23.6\% for
memory alone; on AppWorld, memory alone retains most of the full agent's
performance. These ablations show that some improvements persist without
the retained context, while others depend more strongly on its presence.

\FloatBarrier
\clearpage
\subsection{Transfer across worlds}
\label{app:transfer-crossworld}

\begin{figure}[!htbp]
\centering
\includegraphics[width=\linewidth]{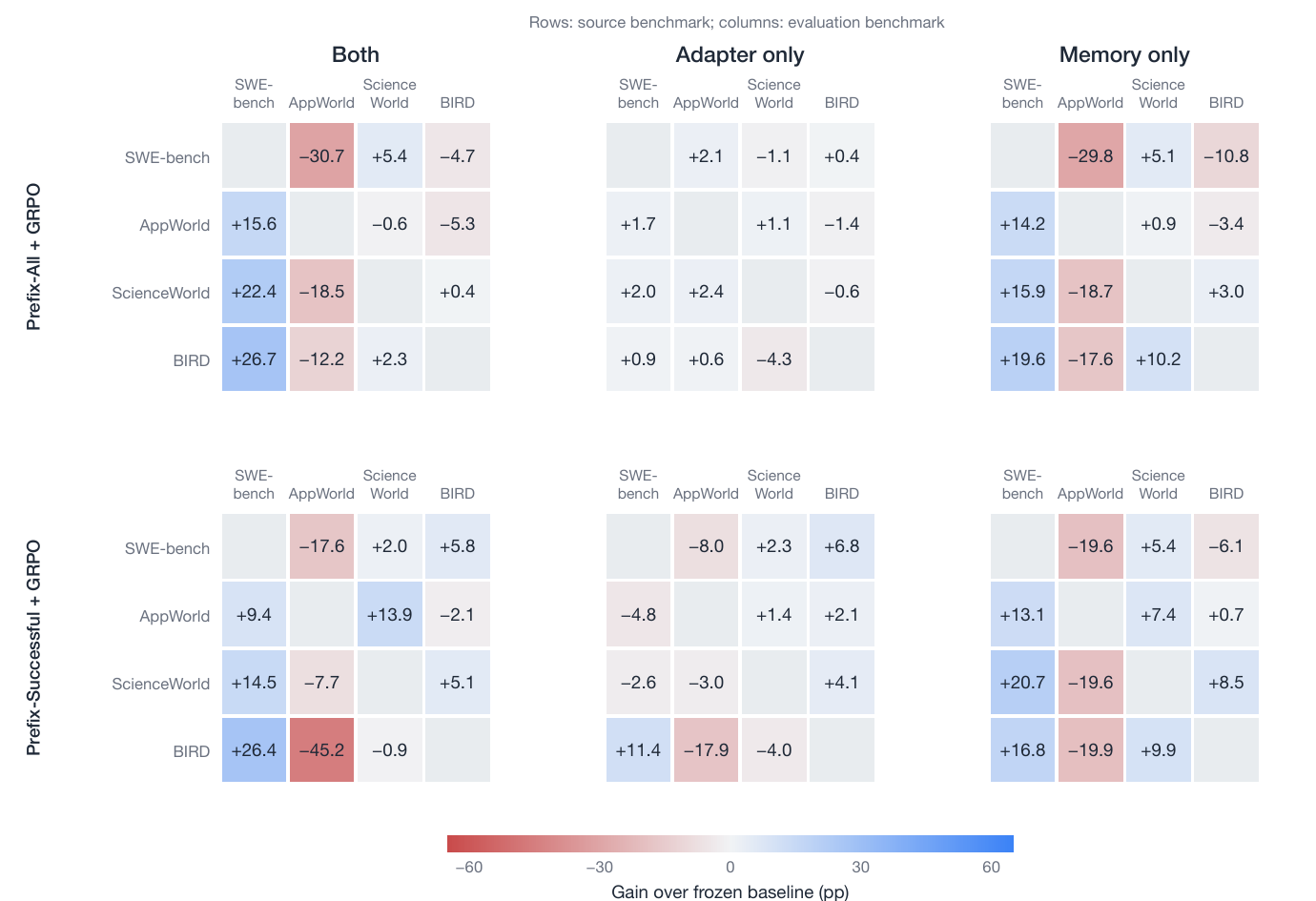}
\caption{\textbf{Transfer of hybrid agents and their components across worlds.}
Within each matrix, rows identify the source benchmark and columns the
evaluation benchmark. Values are success-rate gains in percentage points
over the corresponding round's frozen baseline. Diagonal cells are masked
because they evaluate previously encountered tasks. Both, adapter only,
and memory only use components of the same final seed-4 hybrid agent.}
\label{fig:transfer-crossworld}
\end{figure}

Figure~\ref{fig:transfer-crossworld} evaluates the same agents and their
components on the other worlds. The diagonal is masked because it revisits
tasks used during adaptation. Off-diagonal entries report gains over the
fresh frozen baseline for the target world in the corresponding round.

Transfer depends strongly on the target world. Under both policies,
retained memory from each of the other three worlds improves SWE-bench
success but reduces AppWorld success. The full hybrids show the same
direction of transfer to these two targets. Adapter-only transfer is
mixed and does not reproduce the uniformly positive transfer to
SWE-bench seen with memory. Thus, retained experience can benefit tasks
outside its source environment, but can also hurt performance when moved
to another environment. These results do not identify which aspects of
the transferred behavior cause the changes in success.
\FloatBarrier

\applefootnote{ \textcolor{textgray}{\sffamily Apple and the Apple logo are trademarks of Apple Inc., registered in the U.S. and other countries and regions.}}

\end{document}